\documentclass[journal,twoside,web]{ieeecolor}

\usepackage{tmi}
\usepackage{array}
\usepackage{cite}
\usepackage{extarrows}
\usepackage{mdwmath}
\usepackage{graphicx}         
\usepackage{subfigure}
\usepackage{pifont}
\newcommand{\xmark}{\ding{55}} 
\newcommand{\cmark}{\ding{51}} 
\usepackage{bbm}
\usepackage{dsfont}
\usepackage{bm}
\usepackage{comment}
\usepackage{amsmath}
\usepackage{enumerate}  
\usepackage{mathrsfs}
\usepackage{makeidx}         
\usepackage{graphicx}        
\usepackage{subfigure}
\usepackage{epsfig,amssymb,latexsym}
\usepackage{psfrag}
\usepackage{algorithm}
\usepackage{algorithmicx}
\usepackage{subfigure}
\usepackage{algcompatible,lipsum}
\usepackage{cases}
\usepackage[misc]{ifsym}
\usepackage{fancyhdr}
\usepackage{multirow}
\usepackage{makecell}
\usepackage{color}
\usepackage{indentfirst}
\usepackage[breaklinks=true,bookmarks=false,colorlinks,citecolor=blue,linkcolor=blue]{hyperref}

\usepackage{booktabs}
\usepackage{threeparttable}
\usepackage{siunitx}
\usepackage{tabularx}
\usepackage{colortbl}

\newcommand{\mr}{\mathrm}

\newenvironment{myitemize}{\begin{list}{$\bullet$}
		{\setlength{\topsep}{1mm}
			\setlength{\itemsep}{0.1mm}
			\setlength{\parsep}{0.0mm}
			\setlength{\itemindent}{0mm}
			\setlength{\partopsep}{0mm}
			\setlength{\labelwidth}{1mm}
			\setlength{\leftmargin}{4mm}}}{\end{list}}

\def\BibTeX{{\rm B\kern-.05em{\sc i\kern-.025em b}\kern-.08em
    T\kern-.1667em\lower.7ex\hbox{E}\kern-.125emX}}
\begin{document}
\title{Prototype Correlation
	Matching and Class-Relation Reasoning for Few-Shot Medical Image Segmentation}
\author{Yumin Zhang$^\dag$, 
Hongliu Li$^{\dag, *}$, Yajun Gao, Haoran Duan, Yawen Huang, Yefeng Zheng, \IEEEmembership{Fellow, IEEE}
\thanks{Yumin Zhang and Hongliu Li are with the Department of Civil and Environmental Engineering, The Hong Kong Polytechnic University, Hong Kong, China (e-mail: zhymin456@gmail.com, hongliuli1994@gmail.com). }
\thanks{Yajun Gao is with the Shenyang Institute of Automation, Chinese Academy of Sciences, Shenyang, China (e-mail: gaoyajun@sia.cn).}
\thanks{Haoran Duan is with the Department of Computer Science, Durham University (e-mail: haoran.duan@ieee.org).}
\thanks{Yawen Huang and Yefeng Zheng are with the Jarvis Research Center, Tencent YouTu Lab, Shenzhen, China (e-mail: yawenhuang@tencent.com; yefengzheng@tencent.com).}
\thanks{$^{\dag}$Equal Contributions. $^{*}$The corresponding author is Dr. Hongliu Li.}
}

\maketitle

\begin{abstract}
Few-shot medical image segmentation has achieved great progress in improving accuracy and efficiency of medical analysis in the biomedical imaging field. However, most existing methods cannot explore inter-class relations among base and novel medical classes to reason unseen novel classes. Moreover, the same kind of medical class has large intra-class variations brought by diverse appearances, shapes and scales, thus causing ambiguous visual characterization to degrade generalization performance of these existing methods on unseen novel classes. To address the above challenges, in this paper, we propose a \underline{\textbf{P}}rototype correlation \underline{\textbf{M}}atching and \underline{\textbf{C}}lass-relation \underline{\textbf{R}}easoning (i.e., \textbf{PMCR}) model. The proposed model can effectively mitigate false pixel correlation matches caused by large intra-class variations while reasoning inter-class relations among different medical classes. 
Specifically, in order to address false pixel correlation match brought by large intra-class variations, we propose a prototype correlation matching module to mine representative prototypes that can characterize diverse visual information of different appearances well. We aim to explore prototype-level rather than pixel-level correlation matching between support and query features via optimal transport algorithm to tackle false matches caused by intra-class variations.   
Meanwhile, in order to explore inter-class relations, we design a class-relation reasoning module to segment unseen novel medical objects via reasoning inter-class relations between base and novel classes. Such inter-class relations can be well propagated to semantic encoding of local query features to improve few-shot segmentation performance.
Quantitative comparisons illustrates the large performance improvement of our model over other baseline methods. 
\end{abstract}

\begin{IEEEkeywords}
Medical image segmentation, few-shot segmentation, intra-class variations, inter-class relations. 
\end{IEEEkeywords}
 


\section{Introduction}

\IEEEPARstart{D}{eep} learning-based medical diagnosis methods \cite{9389742, li2018h, 9165815} require a large number of finely annotated medical data to train the segmentation model of medical objects under a supervised learning manner \cite{10.1093/bmb/57.1.193}. Unfortunately, it is difficult and expensive to collect abundant manually-labeled medical data, since only a few doctors with rich clinical experience can annotate privacy-sensitive medical data. Besides, the pre-trained segmentation model of medical classes may suffer from a large performance drop on novel medical objects that are unseen in the training set and are difficult to collect the corresponding well-labeled annotations (e.g., Alzheimer's disease) \cite{9709261, GU2018220}. That is to say, the pre-trained segmentation model has poor generalization performance on unseen medical objects in clinical diagnosis tasks \cite{9165815, What_Transferred_CVPR2020}. More importantly, the significant differences in imaging protocols and plural medical forms among different patients further weaken generalization of trained model to identify unseen medical classes.

\begin{figure}[t]
	\centering
	\includegraphics[width=1.0\linewidth]
	{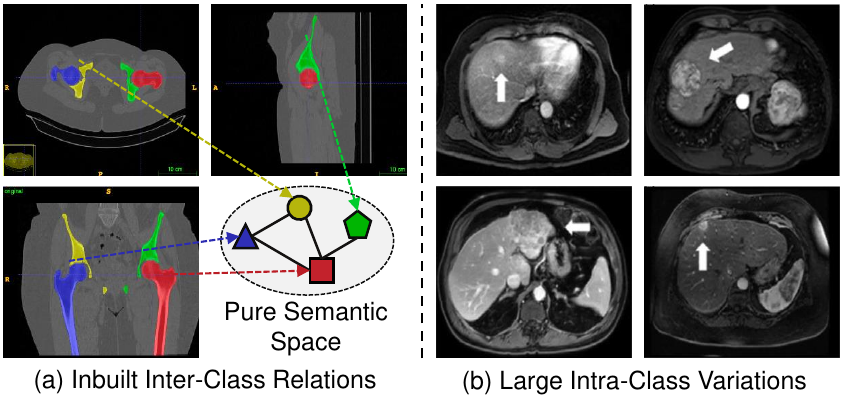}
	\vspace{-20pt}
	\caption{Illustration of two major challenges in the few-shot medical image segmentation. (a) Inbuilt inter-class relations between different medical classes in the pure semantic space. (b) Large intra-class variations within the same medical class under different imaging protocols (e.g., the hepatocellular carcinoma (HCC) has various appearances \cite{10.1093/bib/bbaa295}).  }
	\label{fig:motivation}
\end{figure}

To this end, few-shot semantic segmentation learning \cite{Lang_2022_CVPR, Liu_2022_CVPR, Zhang2021FewShotSV} has become a popular potential solution to improve the generalization performance of the trained model on unseen novel classes when rarely labeled samples of novel classes are available. These methods formulate the few-shot semantic segmentation task as a meta-learning optimization objective \cite{10.5555/3327546.3327622, pmlr-v70-finn17a}. They rely on prototypical learning \cite{Wang_2023_CVPR} or affinity learning \cite{10.1007/978-3-030-58601-0_43} to explore pairwise pixel-level correlation. 
Inspired by the above few-shot methods that are used for natural scene segmentation in the wild \cite{Wang_2023_CVPR, 9616392}, Roy et al. \cite{GUHAROY2020101587} first propose few-shot medical image segmentation task, where the semantic information from support and query samples can be fused to guide semantic segmentation head via the proposed excite and squeeze modules. Following \cite{GUHAROY2020101587}, some works \cite{10.1007/978-3-030-58526-6_45, Huang_2023_CVPR, 9709261} utilize self-supervised superpixel clustering or vector quantization view to perform prototypical learning. Meanwhile, \cite{9711123} uses affinity learning to encode informative contextual characterizations between background and foreground objects while modifying the segmentation mask under an iterative refinement manner.

However, the above few-shot medical image segmentation methods \cite{GUHAROY2020101587, 10.1007/978-3-030-58526-6_45, 9711123, 9709261, Wang_2023_CVPR} neglect \textbf{inter-class relations} among base and novel classes and large \textbf{intra-class variations} within the same medical objects. To understand these challenges in the few-shot medical image segmentation task, as shown in Fig.~\ref{fig:motivation}, we introduce clinical medical diseases analysis as practical examples for better explanations:
\begin{myitemize}
	\item  \textbf{Inter-Class Relations:} Different medical object classes have some kind of inbuilt class relations in the pure semantic space, regardless of the available number of unseen novel medical classes, as shown in Fig.~\ref{fig:motivation}(a). For example, kidney class usually may cause pathology of other related organs such as blood vessels. The inter-class relations between base classes (e.g., kidney) and novel classes (e.g., blood vessels) can be used to recognize unseen medical objects with only few samples. 
	However, existing methods \cite{GUHAROY2020101587, Wang_2023_CVPR} cannot explore intrinsic inter-class relations when segmenting unseen novel medical classes.
\item  \textbf{Intra-Class Variations:} As shown in Fig.~\ref{fig:motivation}(b), the same kind of medical class has large intra-class variations brought by diverse appearances, shapes and scales, thus causing ambiguous visual characterization to degrade generalization performance on unseen novel classes. The hepatocellular carcinoma (HCC) has various appearances under different imaging protocols \cite{10.1093/bib/bbaa295}. 
Besides, the medical diseases from left and right kidneys share similar semantic context but have significantly different visual characterizations (i.e., appearances, scales and shapes) \cite{10.1007/978-3-030-58526-6_45}. 
Such large intra-class variations enforce existing methods \cite{9709261, 9711123} to explore ambiguous visual features, thus causing false match of pairwise pixel correlation among support and query features. This phenomenon can be further amplified by inherent few-shot regimes, decreasing generalization of pre-trained model on unseen abnormal diseases. 	
\end{myitemize}

To address the above challenges, we propose a \textbf{P}rototype correlation \textbf{M}atching and \textbf{C}lass-relation \textbf{R}easoning (i.e., \textbf{PMCR}) model for few-shot medical image segmentation. The proposed PMCR model can effectively mitigate intra-class variations and explore inter-class relations to generalize well on unseen novel medical classes under the optimization manner of meta-learning \cite{10.5555/3327546.3327622}. To be specific, to address false pairwise pixel correlation matches brought by large intra-class variations, we propose a prototype correlation matching module. It first performs singular value decomposition (SVD) on a pixel-wise affinity matrix between support and query features to obtain their representative prototypes. These prototypes can embody diverse intra-class visual information such as different appearances, scales, and shapes. We propagate task contextual information into representative prototypes, while exploring prototype-level rather than pixel-level correlation matching between support and query features via optimal transport algorithm to mitigate intra-class variations. 
Furthermore, considering exploring inter-class relations, we develop a class-relation reasoning module to reason inherent relations among the base and novel classes and then propagate such inter-class relations from support images into the semantic encoding of the query set in the latent feature space. The superpixel centroids generated via foreground superpixel clustering algorithm could well characterize diverse visual properties of different medical classes from the support set. These superpixel centroids of the support set and local features of the query image are then employed to capture inter-class relations via constructing a relation graph. When encoding the query image in the latent feature space, we utilize such intrinsic inter-class relations to modify its original convolution kernel.
We conduct extensive experiments to show the superior performance of our model. We summarize the main contributions of this paper below:


\begin{myitemize}
\item  We develop a new Prototype correlation Matching and Class-relation Reasoning (PMCR) model for few-shot medical image segmentation, which is an earlier work to segment unseen novel medical classes via addressing intra-class variations and exploiting inter-class relations. 
 As we know, it is an earlier work to tackle few-shot medical segmentation from the intra-class and inter-class aspects.

\item Considering false pixel matches caused by large intra-class variations, we propose a prototype correlation matching module to mine representative prototypes, and explore prototype-level rather than pixel-level correlation matching between support and query features via optimal transport.

\item Considering exploring inter-class relations, we develop a class-relation reasoning module to reason inherent relations among base and novel medical classes, and then propagate such inter-class relations from support images into the semantic encoding of query image in latent feature space.
\end{myitemize}

\section{Related Work}

\subsection{Medical Image Segmentation}
Recently, convolution neural networks \cite{7780459, 9616392} have brought widespread attention on biomedical image computing \cite{10.1007/978-3-319-24574-4_28, 7785132, Zhang2021TransFuseFT}, such as registration \cite{8579062, 8633930}, reconstruction \cite{article_CIRCLE, 8416740}, computer-aided diagnosis \cite{9165815} and medical segmentation \cite{What_Transferred_CVPR2020}. 
The success of deep learning-based medical image segmentation has promoted development of diverse applications consisting of anatomical concept understanding \cite{9009566}, tissue segmentation \cite{8811612} and abnormal analysis \cite{li2018h}. 
In the field of medical image segmentation, U-Net \cite{10.1007/978-3-319-24574-4_28} has become one of the widely used frameworks. 
Motivated by U-Net \cite{10.1007/978-3-319-24574-4_28}, diverse variants \cite{7785132, DBLP:journals/corr/abs-1804-03999} are developed to perform medical image segmentation tasks. For example, V-Net \cite{7785132} is designed to tackle class imbalance between background and foreground voxels when processing 3D volume data.  \cite{nnU_Net_2021} focuses on integrating different U-Net \cite{10.1007/978-3-319-24574-4_28} models together via learning optimal configurations for different tasks automatically. 
The above methods \cite{10.1007/978-3-319-24574-4_28, DBLP:journals/corr/abs-1804-03999, GU2018220} require a large number of medical data to train class-specific segmentation model. 
Unfortunately, it is a challenging task to collect abundant finely-labeled medical data in the biomedical analysis field, due to the privacy preservation of patients \cite{Cheng_2022_CVPR}. 

\subsection{Few-Shot Semantic Segmentation}

\begin{figure*}[t]
	\centering
	\includegraphics[width=1.0\linewidth]
	{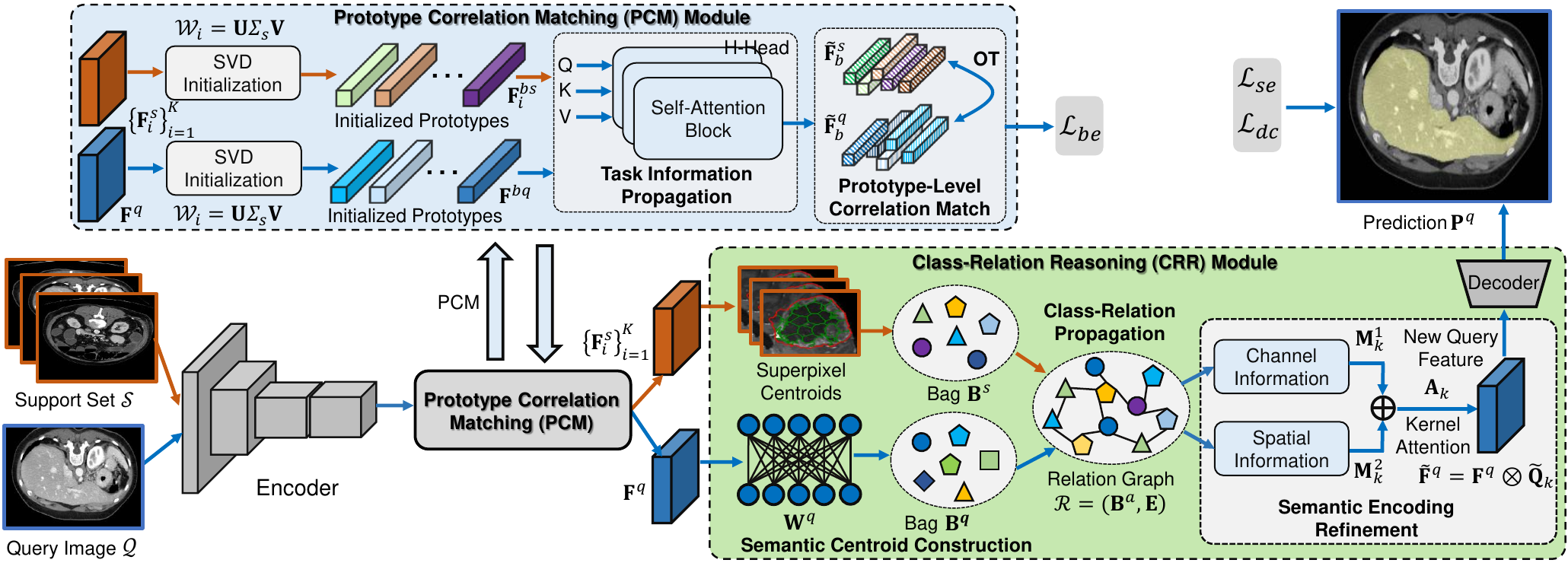}
	\vspace{-20pt}
	\caption{Pipeline of the proposed model. It includes a prototype correlation matching (PCM) module to address false pairwise pixel correlation matches brought by large intra-class variations within the same medical class; and a class-relation reasoning (CRR) module to explore inter-class relations between base and novel classes for semantic encoding of local query features.   }
	\label{fig:overview}
\end{figure*}

For the few-shot medical image segmentation task \cite{DBLP:journals/corr/abs-1810-12241, 10.5555/3294996.3295163, hajimiri2023diam, 10.1145/3386252}, most researches rely on image generation of novel classes to enrich medical data via using few labels \cite{9157721, 8953991}. Unfortunately, they need to retrain the medical diagnosis model when segmenting new medical objects. Roy et al. \cite{GUHAROY2020101587} first apply few-shot learning to the medical image segmentation task, where the semantic information from support and query samples can be fused to guide semantic segmentation head via the proposed excite and squeeze modules. \cite{10.1007/978-3-030-58526-6_45} develops a superpixel clustering-based self-supervised learning framework to explore local informative category-wise prototypes. Tang et al. \cite{9711123} design a context relation encoder to encode informative contextual characterizations between background and foreground regions, while they modify the segmentation mask under an iterative refinement manner. For better generalization performance, Huang et al. \cite{Huang_2023_CVPR} focus on exploring limited-quantity and representative prototypes. 
However, the above methods \cite{10.1007/978-3-030-58526-6_45, GUHAROY2020101587, 8953991, Lang_2022_CVPR} neglect inter-class relations among base and novel classes and large intra-class variations within the same medical class.

\section{The Proposed Model}

\subsection{Problem Definition and Overview}
\textbf{Problem Definition: }
For the conventional few-shot medical image segmentation task \cite{10.1007/978-3-030-58526-6_45, Wang_2019_ICCV}, we use a finely-annotated training set $\mathcal{D}_{tr}$ containing images from training classes $\mathcal{C}_{tr}$ to train the segmentation model. After training, we evaluate the performance of this well-trained segmentation model on a testing set $\mathcal{D}_{te}$ with images from unseen classes $\mathcal{C}_{te}$, where only a few labeled samples of each medical class in $\mathcal{C}_{te}$ are available. There are no overlapping classes between training and test sets (\emph{i.e.}, $\mathcal{C}_{tr}\cap \mathcal{C}_{te} = \varnothing$) in the few-shot medical image segmentation \cite{Lang_2022_CVPR, Liu_2022_CVPR}. Generally, we refer to the few-shot learning problem as $N$-way $K$-shot learning, where $K$ denotes the number of available samples for each medical class in $\mathcal{C}_{te}$, and $N$ is the number of medical classes in $\mathcal{C}_{te}$.

Motivated by baseline few-shot medical image segmentation methods \cite{10.1007/978-3-030-58526-6_45, Wang_2019_ICCV, 9709261, 9711123}, we utilize an episodic training strategy to segment unseen medical classes. In the training phase, we sample an episodic $(\mathcal{S}, \mathcal{Q})$ from the training set $\mathcal{D}_{tr}$. In each episodic training $(\mathcal{S}, \mathcal{Q})$, the support set $\mathcal{S} = \{\mathbf{X}^s_i, \mathbf{Y}^s_i\}_{i=1}^K$ includes $K$ pairs of support image $\mathbf{X}^s_i$ and its corresponding mask $\mathbf{Y}^s_i$ from $\mathcal{D}_{tr}$, while the query set $\mathcal{Q} = \{\mathbf{X}^q, \mathbf{Y}^q\}$ comprises one pair of query image $\mathbf{X}^q$ and its mask $\mathbf{Y}^q$ from $\mathcal{D}_{tr}$. We denote the width and height of images $\{\mathbf{X}_i^s, \mathbf{X}^q\}$ as $W$ and $H$, respectively. With this episodic training strategy, the segmentation model is trained to distill class-contextual correlations from the support set $\mathcal{S} = \{\mathbf{X}^s_i, \mathbf{Y}^s_i\}_{i=1}^K$, and rely on such correlations to make pixel-level predictions on query image $\mathbf{X}^q$. In the testing phase, we also construct an episodic $(\mathcal{S}, \mathcal{Q})$ from the test set $\mathcal{D}_{te}$, where the support set $\mathcal{S} = \{\mathbf{X}^s_i, \mathbf{Y}^s_i\}_{i=1}^K$ containing $K$-shot images of unseen classes $\mathcal{C}_{te}$ is sampled from $\mathcal{D}_{te}$, and the query set $\mathcal{Q} = \{\mathbf{X}^q\}$ only contains the query image $\mathbf{X}^q$ from $\mathcal{D}_{te}$. In the few-shot medical image segmentation task \cite{10.1007/978-3-030-58526-6_45, 9711123}, the trained segmentation model aims to segment unseen novel medical classes of query image $\mathbf{X}^q$.

\textbf{Overview: }
The overview of proposed model is shown in Fig.~\ref{fig:overview}. Our proposed model belongs to the branch of prototypical learning in the few-shot medical image segmentation \cite{Wang_2019_ICCV, 9709261}, where we use the class prototypes of support set to guide the segmentation of query images. 
After extracting latent features $\{\mathbf{F}_i^s\}_{i=1}^K$ and $\mathbf{F}^q$ of the medical support set $\mathcal{S}$ and query set $\mathcal{Q}$, we first forward them into the prototype correlation matching (PCM) module in Section \ref{sec: intra_class} to address false pixel matches between the query and support features brought by large intra-class variation. The PCM module can perform prototype-level correlation match to tackle these false pixel matches via mining some representative prototypes of support and query sets. Then the latent features $\{\mathbf{F}_i^s\}_{i=1}^K$ and $\mathbf{F}^q$ are fed into the class-relation reasoning (CRR) module in Section \ref{sec: inter_class} to explore inter-class relations between base and novel medical classes. We then use such inter-class relations to refine the convolution kernel of encoding local query features and obtain the final query feature. Finally, we input the final query feature into the decoder to perform pixel-level predictions.

\subsection{Prototype Correlation Matching} \label{sec: intra_class}

In the few-shot image segmentation task, the presence of large intra-class variations within the same medical class enforces existing few-shot medical image segmentation methods \cite{10.1007/978-3-030-58526-6_45, 9709261, 9711123} to capture ambiguous visual characteristics. This can lead to the mismatches in pairwise pixel correlations between query and support features. To make matters worse, few-shot regimes can further amplify negative impact of false pairwise pixel correlation matches, leading to sub-optimal results for few-shot medical image segmentation. 
As shown in Fig.~\ref{fig:overview}, after forwarding $K$ support images $\{\mathbf{X}^s_i, \mathbf{Y}_i^s\}_{i=1}^K$ and query image $\mathbf{X}^q$ to the encoder (feature extractor), we can obtain $K$ support features $\{\mathbf{F}_i^s\in\mathbb{R}^{D_l\times H_l\times W_l}\}_{i=1}^K$ and query feature $\mathbf{F}^q\in\mathbb{R}^{D_l\times H_l\times W_l}$. Then we use the mask $\mathbf{Y}_i^s$ to extract the foreground pixel features of support images and neglect noisy background information. It can transform the original support features $\{\mathbf{F}_i^s\in\mathbb{R}^{D_l\times H_l\times W_l}\}_{i=1}^K$ to $\{\tilde{\mathbf{F}}_i^s\in\mathbb{R}^{D_l\times N_i}\}_{i=1}^K$, where $N_i$ denotes the number of foreground pixels inside the mask $\mathbf{Y}_i^s$, $D_l, H_l$ and $W_l$ denote the channel, height and width of latent features.

After we reshape the original query feature $\mathbf{F}^q$ to $\tilde{\mathbf{F}}^q\in\mathbb{R}^{D_l\times H_lW_l}$. The similarity matrix $\mathcal{W}_i\in\mathbb{R}^{H_lW_l\times N_i}$ between the query feature $\tilde{\mathbf{F}}^q$ and the $i$-th support feature $\tilde{\mathbf{F}}_i^s$ is formulated as $\mathcal{W}_i = (\tilde{\mathbf{F}}^q)^{\top} \tilde{\mathbf{F}}_i^s$, where $\{\mathcal{W}_i\}_{i=1}^K$ can measure pairwise pixel correlations between support and query features, which is commonly used by exiting few-shot medical image segmentation methods \cite{10.1007/978-3-030-58526-6_45, 9709261, Huang_2023_CVPR, 9711123} to segment unseen medical classes. However, as mentioned above, $\{\mathcal{W}_i\}_{i=1}^K$ have some false pixel correlations matches caused by large intra-class variations, which can degrade the segmentation performance of existing few-shot methods \cite{10.1007/978-3-030-58526-6_45, 9709261, Huang_2023_CVPR} on unseen medical classes. 
To address such false pixel matches, we propose a prototype correlation matching module in this paper. As shown in Fig.~\ref{fig:overview}, we first perform singular value decomposition (SVD) on pairwise pixel affinity matrices $\{\mathcal{W}_i\}_{i=1}^K$ between support and query features to initialize their representative prototypes. 
We then enrich these prototypes with task-specific contextual information, and rely on the optimal transport (OT) algorithm to replace traditional pairwise pixel match with prototype-level correlation match between support and query features.

$\bullet$ \textbf{Representative Prototypes Initialization:}
In order to enable representative prototypes to well describe diverse semantic clues of medical query and support features, we perform efficient singular value decomposition (SVD) algorithm \cite{article_Halko2011} on original pairwise pixel affinity matrices $\{\mathcal{W}_i \in\mathbb{R}^{H_lW_l\times N_i}\}_{i=1}^K$, and select the largest $S$ singular values. The SVD formulation of $\mathcal{W}_i$ is written as:
\begin{align}
	\mathcal{W}_i \xlongequal[]{\mr{SVD}} \mathbf{U}\Sigma_s\mathbf{V},
	\label{eq: similarity_SVD}
\end{align}
where $\mathbf{U}\in\mathbb{R}^{H_lW_l\times S}, \Sigma_s\in\mathbb{R}^{S\times S}, \mathbf{V}\in\mathbb{R}^{S\times N_i}$. $\Sigma_s$ is diagonal matrix that contains the largest $S$ singular values to preserve enough pairwise pixel correlation from $\mathcal{W}_i$. The left singular matrix $\mathbf{U}$ can be seen as $S$ orthogonal feature bases in the space of support feature number, and we map the query feature $\tilde{\mathbf{F}}^q\in\mathbb{R}^{D_l\times H_lW_l}$ on these orthogonal feature bases to initialize representative prototypes. 
Likewise, we conduct the same operation to project the $i$-th support feature $\tilde{\mathbf{F}}_i^s\in\mathbb{R}^{D_l\times N_i}$ onto its feature bases $\mathbf{V}$. As a result, we can obtain some representative prototypes for medical support and query images via the following objective: 
\begin{align}	
	\mathbf{F}^{bq} = \mathbf{U}^\top(\mathbf{F}^q)^\top, \quad \mathbf{F}_i^{bs} = \mathbf{V}(\mathbf{F}_i^s)^\top,
	\label{eq: feature_base_initialize}
\end{align}
where $\mathbf{F}^{bq}, \mathbf{F}_i^{bs}\in\mathbb{R}^{S\times D_l}$ denote well-initialized representative prototypes which contain most of the pairwise pixel correlations in $\mathcal{W}_i$. These well-initialized prototypes can also speed up the model convergence during the training stage. In this paper, the number of foreground pixels $N_l$ is much smaller than $H_lW_l$: $N_l \ll H_lW_l$, where $H_l=W_l=32$. Besides, $\mathbf{F}^q$ and $\mathbf{F}_i^s$ have a very small feature dimension ($D_l=256$). In light this, following efficient SVD algorithm \cite{article_Halko2011}, we can efficiently obtain $\mathbf{U}, \Sigma_s$ and $\mathbf{V}$ under a single GPU.

$\bullet$ \textbf{Task Information Propagation:}
To prevent the negative influence of noisy prototypes, we propagate task-specific semantic information into representative prototypes. It encourages prototypes to be more representative and robust to characterize diverse semantic clues from query and support images. Specifically, as shown in Fig.~\ref{fig:overview}, given well-initialized prototypes $\mathbf{F}_i^{bs}$ of the $i$-th support image, we condense contextual information from support feature $\tilde{\mathbf{F}}_i^{s}$ into $\mathbf{F}_i^{bs}$ via multi-head self-attention. As a result, we formulate attention result $\mathbf{A}_s^h\in\mathbb{R}^{S\times d}$ of the $h$-th ($h=1, \cdots, H$) head as follows: 
\begin{align}	
	\mathbf{A}_s^h = \sigma(\frac{\mathbf{F}_i^{bs}\mathbf{W}_q^s((\tilde{ \mathbf{F}}_i^s)^\top\mathbf{W}_k^s)^\top}{\sqrt{d}}) (\tilde{ \mathbf{F}}_i^s)^\top\mathbf{W}_v^s, 
	\label{eq: feature_base_support}
\end{align}
where $\mathbf{W}_q^s, \mathbf{W}_k^s, \mathbf{W}_v^s \in\mathbb{R}^{D_l\times d}$ denote mapping matrices, $d$ is feature dimension, $H$ is the number of heads, and $\sigma$ denotes the sigmoid activation function. We then concatenate the attention results $\{\mathbf{A}_s^h\}_{h=1}^H$ from $H$ heads along feature dimension to obtain new representative prototypes  $\mathbf{A}_s\in\mathbb{R}^{S\times D_l} (D_l=dH)$ of the $i$-th support feature. Likewise, given the well-initialized $\mathbf{F}^{bq}$ and query feature $\tilde{\mathbf{F}}^q$, we can also obtain its attention result $\mathbf{A}_q^h\in\mathbb{R}^{S\times d}$ of the $h$-head:
\begin{align}	
	\mathbf{A}_q^h = \sigma(\frac{\mathbf{F}^{bq}\mathbf{W}_q^q((\tilde{ \mathbf{F}}^q)^\top\mathbf{W}_k^q)^\top}{\sqrt{d}}) (\tilde{ \mathbf{F}}^q)^\top\mathbf{W}_v^q, 
	\label{eq: feature_base_query}
\end{align}
where $\mathbf{W}_q^q, \mathbf{W}_k^q, \mathbf{W}_v^q\in\mathbb{R}^{D_l\times d}$ are projection matrices. After performing multi-head self-attention to obtain $\{\mathbf{A}_q^h\}_{h=1}^H$, we concatenate them together along feature dimension, and denote this concatenated result as new representative prototypes $\mathbf{A}_q\in\mathbb{R}^{S\times D_l}$ of query features.

Obviously, $\mathbf{A}_s$ and $\mathbf{A}_q$ are representative enough to retain abundant contextual information from support and query features, respectively. However, they often have significant distribution gaps, due to the negative influence of cluttered backgrounds and large intra-class variations. To bridge this gap, we aggregate contextual information between $\mathbf{A}_s$ and $\mathbf{A}_q$ to capture the co-occurring foreground objects. Specifically, we first concatenate $\mathbf{A}_s$ and $\mathbf{A}_q$ together to get $\mathbf{F}_b\in\mathbb{R}^{2S\times D_l}$, and then perform multi-head self-attention on $\mathbf{F}_{b}$ to propagate semantic information between different prototypes. Therefore, for the $h$-th ($h=1, \cdots, H$) head, we formulate the attention result $\mathbf{A}_b^h\in\mathbb{R}^{2S\times d}$ of $\mathbf{F}_{b}$ as follows:
\begin{align}
	\mathbf{A}_b^h = \sigma(\frac{\mathbf{F}_b\mathbf{W}_q^b(\mathbf{F}_b\mathbf{W}_k^b)^\top}{\sqrt{d}}) \mathbf{F}_b\mathbf{W}_v^b, 
	\label{eq: feature_base_self}
\end{align}
where $\mathbf{W}_q^b, \mathbf{W}_k^b, \mathbf{W}_v^b \in\mathbb{R}^{D_l\times d}$ are linear projection matrices. Given the attention features $\{\mathbf{A}_b^h\}_{h=1}^H$ from $H$ heads, we concatenate them together as $\tilde{\mathbf{F}}_b\in\mathbb{R}^{2S\times D_l} (D_l=dH)$. In this way, the representative prototypes are amended via aggregating complementary contextual information, and are more representative for the few-shot medical image segmentation task. We then split $\tilde{\mathbf{F}}_b\in\mathbb{R}^{2S\times D_l}$ as support prototype $\tilde{\mathbf{F}}_b^s\in\mathbb{R}^{S\times D_l}$ and query prototype $\tilde{\mathbf{F}}_b^q\in\mathbb{R}^{S\times D_l}$.

$\bullet$ \textbf{Prototype-Level Correlation Match:}
In order to mitigate large intra-class variations, it is intuitive to explore prototype-level correlations match between the support prototype $\tilde{\mathbf{F}}_b^s$ and query prototype $\tilde{\mathbf{F}}_b^q$ via $L_2$ distance or dot product. However, this solution assumes that pairwise prototypes are usually aligned well in advance. This assumption can be significantly violated, when some real-world factors such as cluttered background, intra-class variations and heavy intensity noise cause large heterogeneity of representative prototypes. Moreover, such solutions may be sub-optimal since they consider pairwise prototype correlations equally. In fact, not all prototypes are beneficial when exploring the prototype-level correlation of a specific pair of pixels. To tackle these issues, we employ an optimal transport (OT) strategy with a special cost matrix to explore prototype-level correlation match between query and support features. Given a specific cost matrix, the OT strategy can explore minimum transmission cost between distributions of support and query prototypes, thus measuring accurate and optimal prototype-level matching correlations.

Specifically, we define the cost matrix as $(1-\mathbf{M}_b)\in\mathbb{R}^{S\times S}$, where $\mathbf{M}_b\in\mathbb{R}^{S\times S}=\tilde{\mathbf{F}}_b^s(\tilde{\mathbf{F}}_b^q)^\top$ denotes semantic-similarity matrix between support prototype $\tilde{\mathbf{F}}_b^s\in\mathbb{R}^{S\times D_l}$ and query prototype $\tilde{\mathbf{F}}_b^q\in\mathbb{R}^{S\times D_l}$. The transport matrix is formulated as $\mathbf{T}\in\mathbb{R}^{S\times S}$ and we define the optimization objective as:
\begin{align}	
	\min_{\mathbf{T}\in\mathcal{T}} \mathcal{L}_{ot} = \mr{Tr}(\mathbf{T}^\top(1-\mathbf{M}_b)) + \mu\phi(\mathbf{T}), 
	\label{eq: optimization_OT}
\end{align}
where $\phi(\mathbf{T})=-\sum_{ij}\mathbf{T}_{ij} \log \mathbf{T}_{ij}$ is the entropy regularizer. We empirically set $\mu=0.1$ to control smoothness of transport matrix $\mathbf{T}$. The transport matrix $\mathbf{T}$ has the following constrain:
\begin{align}	
	\mathcal{T}=\{\mathbf{T}\in\mathbb{R}^{S\times S}|\mathbf{T}1=\mathbf{u}, \mathbf{T}^\top1=\mathbf{v} \}, 
	\label{eq: definition_T}
\end{align}
where $\mathbf{u}=\mathcal{N}(\mathbf{F}_i^{bs}\tilde{\mathbf{F}}_i^{s}) \in\mathbb{R}^S$ is normalized similarity distribution between the $i$-th support feature $\tilde{\mathbf{F}}_i^{s}$ and their representative prototype $\mathbf{F}_i^{bs}$; 
$\mathbf{v}=\mathcal{N}(\mathbf{F}^{bq}\tilde{\mathbf{F}}^{q})\in\mathbb{R}^S$ denotes normalized similarity distribution between query feature $\tilde{\mathbf{F}}^{q}$ and their representative prototype $\mathbf{F}^{bq}$; $\mathcal{N}$ is normalization function, enforcing $\mathbf{u}$ and $\mathbf{v}$ to satisfy $\sum_i\mathbf{u}_i=1, \sum_j\mathbf{v}_j=1$.

After several Sinkhorn iterations \cite{NIPS2013_af21d0c9} to optimize Eq.~\eqref{eq: optimization_OT}, we can obtain the optimal transportation matrix $\mathbf{T}^*\in\mathbb{R}^{S\times S}$ effectively, and define prototype-level correlation match as:
\begin{align}
	\mathcal{W}_i^* = \mathbf{M}_b\odot\mathbf{T}^*,
	\label{eq: OT_based_similarity}
\end{align}
where $\odot$ denotes the Hardmard product. In order to prevent representative prototypes from degradation, we design a prototype enhancement loss $\mathcal{L}_{be}$ to encourage prototypes to be more representative to describe prototype-level correlation match:
\begin{align}
	\mathcal{L}_{be} = \sum_i^K\|\mathcal{W}_i^*-\mathcal{W}_i^r\|^2_F,
	\label{eq: feature_enhancement}
\end{align}
where $\mathcal{W}_i^r=\mathbf{F}_i^{bs}(\mathbf{F}^{bq})^\top \in\mathbb{R}^{S\times S}$ denotes similarity between the $i$-th support prototypes $\mathbf{F}_i^{bs}$ and query prototypes $\mathbf{F}^{bq}$. This loss ensures that the co-occurring foreground objects of support and query images have small intra-class variations by performing accurate prototype-level correlation matches. It can significantly improve the segmentation performance of the proposed model on unseen medical classes.

\subsection{Class-Relation Reasoning Module}\label{sec: inter_class}
In the few-shot medical image segmentation \cite{Wang_2019_ICCV, Huang_2023_CVPR, 9709261}, the lack of abundant samples of novel medical classes may cause semantic ambiguity, thus decreasing the generalization performance on novel medical classes. However, different medical classes may have intrinsic inter-class relations in the pure semantic space, regardless of the number of available novel medical classes. In light of this, we argue that inter-class relations between base and novel classes play an important role in improving generalization performance on unseen medical classes. To this end, as shown in Fig.~\ref{fig:overview}, we propose a class-relation reasoning (CRR) module to reason inter-class relations between base and novel medical classes. Specifically, we first use the superpixel clustering algorithm SLIC \cite{6205760} to obtain some superpixel centroids. These superpixel centroids can thoroughly represent diverse visual properties of different classes from the support set. Then we construct a relation graph between superpixel centroids of support set and local features of query image to explore inter-class relations between base and novel medical classes, while we use such inter-class relations to refine semantic encoding of local query features from channel and spatial information perspectives.

$\bullet$ \textbf{Semantic Centroids Construction:} 
After obtaining $K$ support features $\{\mathbf{F}_i^s\in\mathbb{R}^{D_l\times H_l\times W_l}\}_{i=1}^K$ and query feature $\mathbf{F}^q\in\mathbb{R}^{D_l\times H_l\times W_l}$, as shown in Fig.~\ref{fig:overview}, we only perform the superpixel clustering algorithm SLIC \cite{6205760} within the foreground pixel features $\{\tilde{\mathbf{F}}_i^s\in\mathbb{R}^{D_l\times N_i}\}_{i=1}^K$. Specifically, given $N_s$ initial superpixel seeds $\mathbf{S}_i^0\in\mathbb{R}^{D_l\times N_s}$, we can employ an iterative clustering strategy \cite{6205760} to obtain final superpixel centroids. For the $t$-th iteration, a pixel affinity map $\mathbf{Z}_i\in\mathbb{R}^{N_i\times N_s}$ is computed via measuring differentiable similarity between all foreground pixels and superpixel centroids within the $i$-th medical support image. Consequently, the affinity between the $p$-th ($p = 1, \cdots, N_i$) pixel feature $(\tilde{\mathbf{F}}_i^s)_p\in\mathbb{R}^{D_l}$ of $\tilde{\mathbf{F}}_i^s$ and the $f$-th ($f = 1, \cdots, N_s$) superpixel centroid $(\mathbf{S}_i^{t-1})_f \in\mathbb{R}^{D_l}$ of the $(t\!-\!1)$-th iteration is defined as $Z_i^{p, f}$:  
\begin{align}
	Z_i^{p, f} = \exp^{-\Phi((\tilde{\mathbf{F}}_i^s)_p, (\mathbf{S}_i^{t-1})_f)}, 
	\label{eq: affinity_relation} 
\end{align}
where $	Z_i^{p, f}$ is the $(p, f)$-th element of the pixel affinity map $\mathbf{Z}_i$. $\Phi((\tilde{\mathbf{F}}_i^s)_p, (\mathbf{S}_i^{t-1})_f) = \|(\tilde{\mathbf{F}}_i^s)_p -  (\mathbf{S}_i^{t-1})_f\|_F^2$ represents the distance measure function. Then we utilize the affinity map $\mathbf{Z}_i$ to reweight foreground pixel features $\tilde{\mathbf{F}}_i^s$ and obtain new updated superpixel centroid $(\mathbf{S}_i^t)_f$ at the $t$-th iteration:
\begin{align}
	(\mathbf{S}_i^t)_f = \frac{1}{G_i^t}\sum_{p=1}^{N_i} Z_i^{p, f} (\tilde{\mathbf{F}}_i^s)_p,
	\label{eq: superpixel_centroids} 
\end{align}
where $(\mathbf{S}_i^t)_f$ denotes the $f$-th superpixel centroid of $\mathbf{S}_i^t$, and $G_i^t = \sum_{p=1}^{N_i} Z_i^{p, f}$.
The final optimal superpixel centroids of the $i$-th support image are denoted as $\mathbf{S}_i\in\mathbb{R}^{D_l\times N_s}$. Inspired by MaskSLIC \cite{DBLP:journals/corr/Irving16}, we uniformly initialize superpixel seeds within masked foregrounds for fast convergence. 

$\bullet$ \textbf{Class-Relation Propagation:}
The superpixel centroids $\{\mathbf{S}_i\in\mathbb{R}^{D_l\times N_s}\}_{i=1}^K$ can effectively characterize a wide range of visual properties of different medical classes. In light of this, we can use them to explore inbuilt inter-class relations among support and query sets via constructing a relation graph, and employ such inter-class relations to refine semantic encoding of local query features.
Given the superpixel centroids $\{\mathbf{S}_i\}_{i=1}^K$ of $K$ medical support images, we concatenate them together as a support bag
$\mathbf{B}^s = [\mathbf{S}_1, \mathbf{S}_2, \cdots; \mathbf{S}_K] \in\mathbb{R}^{D_l\times N_sK}$. To explore inter-class relations of all categories, we construct a negligible memory $\mathcal{M}$ to randomly store $\nu=5$ superpxiel centroids for each class in the superpxiel generation process. Considering data augmentation, we randomly select some newly-generated superpixel centroids of each class within the support set to update $\mathcal{M}$. If $\mathbf{B}^s$ doesn't contain the superpixel centroids belonging to the $c$-th class, we will pick $\nu$ centroids of the $c$-th class from $\mathcal{M}$ and place them into $\mathbf{B}^s$ to maintain class-balanced semantic integrity. In light of this, the support bag is reformulated as
$\mathbf{B}^s = [\mathbf{S}_1, \cdots, \mathbf{S}_K, \mathbf{S}_{\mathcal{M}}] \in\mathbb{R}^{D_l\times (N_sK+V)}$, where $\mathbf{S}_{\mathcal{M}}$ contains $V$ superpixel centroids sampled from $\mathcal{M}$. 
Here, we can argue that $\mathbf{B}^s$ is composed of $N_sK+V$ superpixel centroids containing the full semantic information of all classes from the support set. 
Considering spatial information of the query feature $\mathbf{F}^q\in\mathbb{R}^{D_l\times H_l\times W_l}$, we then apply a linear layer $\mathbf{W}^q\in\mathbb{R}^{H_l\times W_l\times N_q}$ to transform each channel of $\mathbf{F}^q$ to a global descriptor $\mathbf{B}^q = \mathbf{F}^q\mathbf{W}^q\in\mathbb{R}^{D_l\times N_q}$, where $N_q$ is the number of representative global descriptors. We rewrite $\mathbf{B}^q$ as $[\mathbf{f}_1^q, \mathbf{f}_2^q, \cdots, \mathbf{f}_{N_q}^q]\in\mathbb{R}^{D_l\times N_q}$, and consider it as a query bag to fully describe diverse visual information of the query set.


To explore inter-class relations among different medical classes, we merge the support bag $\mathbf{B}^s$ and the query bag $\mathbf{B}^q$ to obtain a new bag
$\mathbf{B}^a = [\mathbf{B}^s, \mathbf{B}^q] \in \mathbb{R}^{D_l\times N_a}$, where $N_a=N_sK+V+N_q$ denotes the number of representative prototypes within $\mathbf{B}^a$. Obviously, $\mathbf{B}^a$ contains representative feature descriptors of different medical objects from support and query images. 
The affinity between different descriptors in $\mathbf{B}^a$ indicates intrinsic inter-class contextual relations between different medical classes. In light of this, such inter-class relation propagation between support images $\{\mathbf{X}_i^s\}_{i=1}^K$ and query image $\mathbf{X}^q$ plays an important role in determining accurate object boundary of medical query image. It can significantly improve segmentation performance on unseen novel medical classes in the few-shot medical image segmentation task. 
To this end, we propose a relation graph $\mathcal{R} = (\mathbf{B}^a, \mathbf{E})$, where $\mathbf{B}^a\in\mathbb{R}^{D_l\times N_a}$ represents the graph node set including a total of $N_a$ feature descriptors $\{\mathbf{f}_i\}_{i=1}^{N_a}\in\mathbf{B}^a$, and $\mathbf{E}\in\mathbb{R}^{N_a\times N_a}$ is the graph edge set to describe inter-class relationships between different medical classes. To be specific, we utilize two weight matrices $\mathbf{W}_s^1, \mathbf{W}_s^2\in\mathbb{R}^{D_s\times D_l}$ to project feature descriptors $\{\mathbf{f}_i\}_{i=1}^{N_a}\in\mathbf{B}^a$ into a shared $D_s$-dimension feature space, and compute their pairwise  similarity relation $\mathbf{E}$: 
\begin{align}	
	\mathbf{E} = (\mathbf{W}_s^1\mathbf{B}^a)^{\top}(\mathbf{W}_s^2\mathbf{B}^a).
	\label{eq: similarity_relationship} 
\end{align}

When a graph edge has a higher semantic similarity score, it indicates closer contextual relationships between pairwise feature descriptors. The graph edge set $\mathbf{E}$ can describe inter-class affinity among different support classes. To achieve inter-class relation propagation between support set $\{\mathbf{X}_i^s\}_{i=1}^K$ and query image $\mathbf{X}^q$, we utilize a shallow graph convolution network and a residual connection strategy to construct a new semantic bag $\tilde{\mathbf{B}}^n\in\mathbb{R}^{D_l\times N_a}$:
\begin{align}	
\tilde{\mathbf{B}}^n = \mathbf{B}^a +  \delta((\mathbf{E}(\mathbf{B}^a)^{\top}\mathbf{W}_g)^{\top}) \odot\mathbf{B}^a,
\label{eq: new_contextual_bag}
\end{align}
where $\mathbf{W}_g\in\mathbb{R}^{D_l\times D_l}$ denotes the parameter weights of graph convolution network, $\delta$ is softmax function, and $\odot$ denotes the Hardmard product. The new semantic bag $\tilde{\mathbf{B}}^n$ obtained via Eq.~\eqref{eq: new_contextual_bag} has considered inter-class relation propagation among different medical classes via the relation graph $\mathcal{R}$.

$\bullet$ \textbf{Semantic Encoding Refinement:}
To propagate inter-class relationships within $\tilde{\mathbf{B}}^n$ into the semantic encoding of local query feature $\mathbf{F}^q$, as shown in Fig.~\ref{fig:overview}, we construct a convolution kernel attention $\mathbf{A}_k\in\mathbb{R}^{D_u\times D_l\times n_1\times n_2}$ to refine original convolution kernel $\mathbf{Q}_k\in\mathbb{R}^{D_u\times D_l\times n_1\times n_2}$, where 
$D_u$ denotes output channel of the query feature $\mathbf{F}^q$, $n_1$ and $n_2$ are kernel sizes. 
To reduce computation burden of refining original convolution kernel, we decompose $\mathbf{A}_k$ as two different tensors $\mathbf{M}_k^1\in\mathbb{R}^{D_u\times n_1\times n_2}$ and $\mathbf{M}_k^2\in\mathbb{R}^{D_l\times n_1\times n_2}$. Then we explore both channel and spatial information to refine semantic contextual encoding of local query features.

\textbf{Channel Information:} 
To explore channel information of $\tilde{\mathbf{B}}^n\in\mathbb{R}^{D_l\times N_a}$, we employ a linear layer with weight as $\mathbf{W}_c\in\mathbb{R}^{D_u\times D_l}$ to transform $\tilde{\mathbf{B}}^n$ into a discriminative feature space. Then we can obtain a new feature descriptor $\mathbf{F}^n = \mathbf{W}_c\tilde{\mathbf{B}}^n \in\mathbb{R}^{D_u\times N_a}$, which can  effectively capture channel contextual information from $\tilde{\mathbf{B}}^n$.

\textbf{Spatial Information:}
Meanwhile, we use two linear layers $\mathbf{W}_c^1, \mathbf{W}_c^2 \in\mathbb{R}^{N_a\times n_1\times n_2}$ to encode $\mathbf{F}^n$ and $\tilde{\mathbf{B}}^n\in\mathbb{R}^{D_l\times N_a}$, respectively. Afterwards, we obtain two tensors $\mathbf{M}_k^1\in\mathbb{R}^{D_u\times n_1\times n_2}$ and $\mathbf{M}_k^2\in\mathbb{R}^{D_l\times n_1\times n_2}$ via $\mathbf{M}_k^1 = \mathbf{F}^n\mathbf{W}_c^1$ and $\mathbf{M}_k^2 = \tilde{\mathbf{B}}^n \mathbf{W}_c^2$.
Then we apply an element-wise sum operation among $\mathbf{M}_k^1$ and $\mathbf{M}_k^2$ to compute convolution kernel attention $\mathbf{A}_k = \mathbf{M}_k^1 \oplus\mathbf{M}_k^2\in\mathbb{R}^{D_u\times D_l\times n_1\times n_2}$, where the $(a, b, c, d)$-th element $(\mathbf{A}_k)_{a,b,c,d}$ of $\mathbf{A}_k$ is written as follows:
\begin{align}	
	(\mathbf{A}_k)_{a,b,c,d} = \delta((\mathbf{M}_k^1)_{a, c, d} + (\mathbf{M}_k^2)_{b,c,d}),
	\label{eq: kernel_mask_sum}
\end{align}
where $a,b,c,d$ are indexes of $\mathbf{A}_k$, and $\delta$ is softmax function.

The convolution kernel attention $\mathbf{A}_k$ is utilized to refine original convolution kernel $\mathbf{Q}_k$ and obtain a new convolution kernel $\tilde{\mathbf{Q}}_k = \mathbf{A}_k\odot\mathbf{Q}_k\in\mathbb{R}^{D_u\times D_l\times n_1\times n_2}$, as shown in Fig.~\ref{fig:overview}. Here, $\tilde{\mathbf{Q}}_k$ has considered inter-class relationships between different medical classes from support and query sets. Then we employ $\tilde{\mathbf{Q}}_k$ to refine convolution kernel of encoding local query feature $\mathbf{F}^q$, and obtain the final query feature $\tilde{\mathbf{F}}^q = \mathbf{F}^q\otimes\tilde{\mathbf{Q}}_k$, where $\otimes$ is convolution operation. Thus, the semantic encoding of local query features is supervised via inherent inter-class relations between different medical objects from support and query sets. The proposed class-relation reasoning (CRR) module can explore inter-class among different base classes in the training phase. When we use $K$-shot images of unseen classes to finetune the segmentation model in the testing phase, this CRR module can capture inter-class between base and novel classes via randomly selecting some superpixel centroids form $\mathcal{M}$ to improve segmentation performance on unseen medical classes.

\subsection{Optimization Loss}
In this paper, we use three loss functions to optimize the proposed model: segmentation loss $\mathcal{L}_{\mr{se}}$, Dice loss $\mathcal{L}_{\mr{dc}}$ and a prototype enhancement loss $\mathcal{L}_{be}$ defined in Eq.~\eqref{eq: feature_enhancement}. For the segmentation loss $\mathcal{L}_{\mr{se}}$, we use the adaptive local prototype pooling proposed in \cite{10.1007/978-3-030-58526-6_45} to obtain class prototypes, and compute class-wise similarities between the final query feature $\tilde{\mathbf{F}}^q$ and these class prototypes to obtain a similarity-based classifier. As introduced in \cite{10.1007/978-3-030-58526-6_45}, this classifier can normalize class-wise similarities into probability map $\mathbf{P}^q\in\mathbb{R}^{H\times W\times C}$, where $C$ denotes the number of classes. Given the one-hot groundtruth $\mathbf{G}^q\in\mathbb{R}^{H\times W\times C}$ of the input query image $\mathbf{X}^q$,  the segmentation loss $\mathcal{L}_{\mr{se}}$ is expressed as follows:
\begin{align}
	\mathcal{L}_{\mr{se}} = -\frac{1}{H_oW_oC}\sum_{i=1}^{H_o}\sum_{j=1}^{W_o}\sum_{c=1}^{C}\mathbf{G}^q_{i,j,c} \log(\mathbf{P}^q_{i,j,c}),
	\label{eq: cross_entropy_loss} 
\end{align} 
where $\mathbf{P}^q_{i,j,c}$ and $\mathbf{G}^q_{i,j,c}$ denote the $(i,j,c)$-th element of $\mathbf{P}^q$ and $\mathbf{G}^q$, respectively. 
Besides, we follow the traditional few-shot medical image work \cite{9711123} to define the Dice loss $\mathcal{L}_{\mr{dc}}$ as:
\begin{align}	
	\mathcal{L}_{\mr{dc}} = 1 - \frac{2\sum_{i=1}^{H}\sum_{j=1}^{W}\sum_{c=1}^{C} \mathbf{P}^q_{i,j,c}\mathbf{G}^q_{i,j,c}}{P_{\mr{all}} + G_{\mr{all}}},
	\label{eq: dice_loss} 
\end{align} 
where $P_{\mr{all}} = \sum_{i=1}^{H}\sum_{j=1}^{W}\sum_{c=1}^{C} \mathbf{P}^q_{i,j,c}$ and $G^q_{\mr{all}} = \sum_{i=1}^{H}\sum_{j=1}^{W}\sum_{c=1}^{C}\mathbf{G}^q_{i,j,c}$ represent the sum of $\mathbf{P}^q$ and $\mathbf{G}^q$. In conclusion, the overall optimization objective $\mathcal{L}_{\mr{all}}$ is:
\begin{align}	
	\mathcal{L}_{\mr{all}} = \mathcal{L}_{\mr{se}} + \lambda_1 \mathcal{L}_{be} + \lambda_2\mathcal{L}_{\mr{dc}}, 
	\label{eq: overall_objective} 
\end{align} 
where $\lambda_1,\lambda_2$ are used to control the effect of $\mathcal{L}_{be}$ and $\mathcal{L}_{\mr{dc}}$. In Eq.~\eqref{eq: overall_objective}, we follow the conventional few-shot medical image work \cite{9711123} to set $\lambda_2$ as 1.0, and set $\lambda_1=0.5$ empirically.

\section{Experiments}

\begin{table}[t]
\centering
\setlength{\tabcolsep}{1.5mm}
\caption{Comparisons (Dice score) on the Cardiac-MRI \cite{8458220} dataset under the setting \#1. }
\resizebox{\linewidth}{!}{
\begin{tabular}{l|c|ccc|cl}
\toprule[1pt]
\makecell[c]{Comparison Methods} & SSL & LV-BP & LV-MYO & RV & Avg. & Imp.  \\
\midrule[1pt]
SE-Net \cite{GUHAROY2020101587} & \xmark & 58.04 & 25.18 & 12.86 & 32.03 & +55.04 \\
Vanilla PANet \cite{Wang_2019_ICCV}  & \xmark & 53.64 & 35.72 &39.52 & 42.96 & +44.11 \\ 
ALPNet \cite{10.1007/978-3-030-58526-6_45} & \xmark &73.08 &49.53& 58.50 &60.34 & +26.73 \\
Affine \cite{9711123} & \xmark & 74.82 & 51.43 & 72.89 & 66.38 & +20.69  \\
\midrule
SSL-PANet \cite{Wang_2019_ICCV} & \cmark &70.43& 46.79& 69.52 &62.25 & +24.82 \\
SSL-ALPNet \cite{10.1007/978-3-030-58526-6_45} & \cmark &83.99 &66.74& 79.96 &76.90 & +10.17 \\
RP-Net \cite{9711123} & \cmark & 84.63 & 67.18 & 82.15 & 77.99 & +9.08 \\
NTRE \cite{Liu_2022_CVPR}  & \cmark & 82.17 & 58.25 & 77.06 & 72.49 & +14.58 \\
HDCD \cite{peng2023cvpr}  & \cmark & 85.08 & 68.42 & 83.16 & 78.89 & +8.18 \\
ALPNet-BP \cite{9709261} & \cmark & 83.98 & 67.68 & 82.15 & 77.94 & +9.13 \\
POP \cite{liu2023learning} & \cmark & 82.17 & 67.15& 82.78 & 77.37 & +9.70 \\
SSL-VQ \cite{Huang_2023_CVPR} & \cmark & 89.68 & 78.27 & 86.64 & 84.86 & +2.21 \\
\midrule
\textbf{Ours (PMCR)} & \cmark & \textbf{\textcolor{red}{91.06}} & \textbf{\textcolor{red}{80.84}} & \textbf{\textcolor{red}{89.31}} & \textbf{\textcolor{red}{87.07}} & -- \\
\bottomrule[1pt]
\end{tabular}}
\label{tab: cardiac_setting_1} 
\end{table}

\begin{table}[t]
\centering
\setlength{\tabcolsep}{1.5mm}
\caption{Comparisons (Dice score) on the Abdominal-MRI \cite{KAVUR2021101950} dataset under the setting \#1 (top block) and the setting \#2 (bottom block). }
\resizebox{\linewidth}{!}{
\begin{tabular}{l|c|cccc|cc}
\toprule[1pt]
\makecell[c]{Comparison Methods} & SSL & LK & RK & Spleen & Liver & Avg. & Imp. \\
\midrule[1pt]
SE-Net \cite{GUHAROY2020101587} & \xmark & 45.78 &47.96& 47.30& 29.02& 42.51 & +44.86 \\
Vanilla PANet \cite{Wang_2019_ICCV}& \xmark &30.99& 32.19 &40.58 &50.40& 38.53& +48.84 \\ 
ALPNet \cite{10.1007/978-3-030-58526-6_45}  & \xmark &44.73 &48.42 &49.61& 62.35 &51.28 & +36.09 \\
Affine \cite{9711123}  & \xmark & 67.51 & 64.28 & 61.84 & 71.62 & 66.31 & +21.06 \\
\midrule
SSL-PANet \cite{Wang_2019_ICCV}  & \cmark &58.83 &60.81& 61.32& 71.73 &63.17 & +24.20 \\
SSL-ALPNet \cite{10.1007/978-3-030-58526-6_45} & \cmark &81.92 &85.18& 72.18 &76.10 &78.84& +8.53 \\
RP-Net \cite{9711123}  & \cmark & 83.16 & 83.76 & 73.85 & 78.27 & 79.76 & +7.61 \\
NTRE \cite{Liu_2022_CVPR} & \cmark & 76.39 & 79.54 & 69.82 & 71.16 & 74.23 & +13.14\\
HDCD \cite{peng2023cvpr} & \cmark & 72.17 & 73.45 & 76.04 & 74.37 & 74.00 & +13.37 \\
ALPNet-BP \cite{9709261} & \cmark & 70.21 & 83.26 & 74.38 & 77.04 & 76.22 & +11.15 \\
POP \cite{liu2023learning} & \cmark & 74.33 & 71.45 & 73.68 & 79.61 & 74.77 & +12.6 \\
SSL-VQ \cite{Huang_2023_CVPR} & \cmark & 79.92 & 77.21 & 91.56 & 89.54 & 84.56 & +2.81 \\
\midrule
\textbf{Ours (PMCR)} & \cmark & \textbf{\textcolor{red}{83.27}} & \textbf{\textcolor{red}{84.04}} & \textbf{\textcolor{red}{92.33}} & \textbf{\textcolor{red}{89.82}} & \textbf{\textcolor{red}{87.37}} & -- \\
\midrule[1pt]
\midrule[1pt]
	SE-Net \cite{GUHAROY2020101587} & \xmark & 62.11& 61.32& 51.80& 27.43 &50.66 & +30.87 \\
Vanilla PANet \cite{Wang_2019_ICCV} & \xmark & 53.45& 38.64 &50.90& 42.26& 46.33& +35.20 \\ 
ALPNet \cite{10.1007/978-3-030-58526-6_45} & \xmark & 53.21& 58.99 &52.18 &37.32& 50.43 & +31.10 \\
Affine \cite{9711123} & \xmark & 62.87& 64.70& 69.10& 65.00& 65.41 & +16.12  \\
\midrule
ALPNet-init \cite{10.1007/978-3-030-58526-6_45} & \cmark & 19.28 &14.93 &23.76& 37.73 &23.93& +57.60 \\
SSL-PANet \cite{Wang_2019_ICCV} & \cmark & 47.71& 47.95& 58.73& 64.99& 54.85& +26.68 \\
SSL-ALPNet \cite{10.1007/978-3-030-58526-6_45} & \cmark & 73.63 &78.39& 67.02 &73.05 &73.02& +8.51 \\
RP-Net \cite{9711123} & \cmark & 76.35& 81.40& 85.78& 73.51& 79.26 & +2.27 \\
NTRE \cite{Liu_2022_CVPR} & \cmark & 68.48 & 71.32 & 73.06 & 62.84 & 68.93 & +12.60 \\
HDCD \cite{peng2023cvpr} & \cmark & 76.61 & 79.85 & 82.45 & 71.28 & 77.55 & +3.98 \\
ALPNet-BP \cite{9709261} & \cmark & 78.77 & 83.44 & 70.02 & 75.01 & 76.81 & +4.72 \\
POP \cite{liu2023learning} & \cmark & 74.52 & 78.64 & 71.17 & 73.52 & 74.46 & +7.07 \\
SSL-VQ \cite{Huang_2023_CVPR} & \cmark & 77.24 & 83.15 & 79.61 & 74.40 & 78.60 & +2.93 \\
\midrule
\textbf{Ours (PMCR)} & \cmark & \textbf{\textcolor{red}{78.89}} & \textbf{\textcolor{red}{84.11}} & \textbf{\textcolor{red}{87.64}} & \textbf{\textcolor{red}{75.49}} & \textbf{\textcolor{red}{81.53}} & -- \\

\bottomrule[1pt]

\end{tabular}}
\label{tab: Abdominal_MRI} 
\end{table}

\subsection{Datasets and Evaluation}
We utilize four benchmark medical datasets to verify the effectiveness of the proposed model under different settings of few-shot medical image segmentation: cardiac segmentation for Magnetic Resonance Imaging (MRI) scans (\textbf{Cardiac-MRI} \cite{8458220}), abdominal Computed Tomography (CT) scans\footnote{https://paperswithcode.com/dataset/miccai-2015-multi-atlas-abdomen-labeling} (\textbf{Abdominal-CT} \cite{10.1007/978-3-030-58526-6_45}), abdominal Magnetic Resonance Imaging (MRI) scans (\textbf{Abdominal-MRI} \cite{KAVUR2021101950}) and Prostate Magnetic Resonance Imaging (MRI) scans ((\textbf{Prostate-MRI} \cite{Huang_2023_CVPR}).
Inspired by conventional few-shot medical image segmentation methods \cite{Lang_2022_CVPR, Liu_2022_CVPR, Wang_2019_ICCV, 9709261, GUHAROY2020101587, Huang_2023_CVPR}, we follow the same experimental configurations for fair comparisons. Specifically, we reformulate all the slices of Cardiac-MRI \cite{8458220}, Abdominal-MRI \cite{KAVUR2021101950} and Abdominal-CT \cite{10.1007/978-3-030-58526-6_45} dataset as 2D short-axis slices. Then we resize them as $256\times 256$ pixels, and copy the intensity to all three color channels to exploit the pre-trained networks on natural images. This processed 2D slice is then regarded as the input of the proposed model.

\begin{table}[t]
	\centering
	\setlength{\tabcolsep}{1.5mm}
	\caption{Comparisons (Dice score) on the Abdominal-CT \cite{10.1007/978-3-030-58526-6_45} dataset under the setting \#1 (top block) and the setting \#2 (bottom block). }
\resizebox{\linewidth}{!}{
		\begin{tabular}{l|c|cccc|cl}
			\toprule[1pt]
			\makecell[c]{Comparison Methods} & SSL & LK & RK & Spleen & Liver & Avg. & Imp.\\
			\midrule[1pt]
			SE-Net \cite{GUHAROY2020101587} & \xmark & 24.42 &12.51& 43.66& 35.42& 29.00 & +48.23 \\
			Vanilla PANet \cite{Wang_2019_ICCV} & \xmark & 20.67& 21.19& 36.04& 49.55& 31.86 & +45.37 \\ 
			ALPNet \cite{10.1007/978-3-030-58526-6_45} & \xmark &29.12 &31.32& 41.00 &65.07 &41.63 & +35.60 \\
			Affine \cite{9711123} & \xmark & 64.26 & 65.16 & 59.43 &71.15 & 65.00 & +12.23  \\
			\midrule
			SSL-PANet \cite{Wang_2019_ICCV} & \cmark &56.52 &50.42& 55.72 &60.86& 57.88 & +19.35 \\
			SSL-ALPNet \cite{10.1007/978-3-030-58526-6_45} & \cmark &72.36 &71.81& 70.96& 78.29& 73.35& +3.88 \\
			RP-Net \cite{9711123} & \cmark & 74.33 & 73.06 & 72.18 & 79.96 & 74.88 & +2.35 \\
			NTRE \cite{Liu_2022_CVPR} & \cmark & 71.01 & 68.93 & 67.49 & 74.63 & 70.52 & +6.71 \\
			HDCD \cite{peng2023cvpr} & \cmark & 70.36 & 71.22 & 74.19 & 73.82 & 72.40 & +4.83 \\
			ALPNet-BP \cite{9709261} & \cmark & 74.38 & 72.56 & 72.49 & 79.06 & 74.62 & +2.61 \\
			POP \cite{liu2023learning} & \cmark & 72.16 & 70.03 & 69.94 & 75.18 & 71.83 & +5.40 \\
			SSL-VQ \cite{Huang_2023_CVPR} & \cmark & 75.16 & 73.08 & 73.22 & 80.17 & 75.41 & +1.82 \\
			\midrule
			\textbf{Ours (PMCR)} & \cmark & \textbf{\textcolor{red}{77.16}} & \textbf{\textcolor{red}{75.27}} & \textbf{\textcolor{red}{74.34}} & \textbf{\textcolor{red}{82.16}} & \textbf{\textcolor{red}{77.23}} & -- \\
			
			\midrule[1pt]
			\midrule[1pt]
			
			SE-Net \cite{GUHAROY2020101587} & \xmark & 32.83& 14.34& 0.23& 0.27 & 11.91 & +62.42 \\
			Vanilla PANet \cite{Wang_2019_ICCV} & \xmark & 32.34& 17.37& 29.59& 38.42& 29.43& +44.90 \\ 
			ALPNet \cite{10.1007/978-3-030-58526-6_45} & \xmark & 34.96& 30.40 &27.73 &47.37& 35.11 & +39.22 \\
			Affine \cite{9711123} & \xmark & 48.99& 43.44& 45.67& 68.93& 51.75& +22.58  \\
			\midrule
			ALPNet-init \cite{10.1007/978-3-030-58526-6_45} & \xmark & 13.90& 11.61& 16.39& 41.71 &20.90 & +53.43 \\
			SSL-PANet \cite{Wang_2019_ICCV} & \xmark & 37.58 &34.69& 43.73& 61.71& 44.42& +29.91 \\
			SSL-ALPNet \cite{10.1007/978-3-030-58526-6_45}& \xmark & 63.34& 54.82 &60.25 &73.65& 63.02& +11.31 \\
		
			RP-Net \cite{9711123} & \cmark &69.85& 70.48&70.00& 79.62&72.48& +1.85 \\
			NTRE \cite{Liu_2022_CVPR} & \cmark & 57.33 & 64.26 & 54.68 & 71.13 & 61.85 & +12.48 \\
			HDCD \cite{peng2023cvpr} & \cmark & 67.46 & 68.73 & 70.12 & 80.04 & 71.59 & +2.74 \\
			ALPNet-BP \cite{9709261} & \cmark & 66.04 & 62.14 & 68.39 & 73.90 & 67.62 & +6.71 \\
			POP \cite{liu2023learning} & \cmark & 63.27 & 60.15 & 65.42 & 70.38 & 64.81 & +9.52 \\
			SSL-VQ \cite{Huang_2023_CVPR} & \cmark & 68.18 & 68.73 & 70.13 & 77.82 & 71.22 & +3.11 \\
			\midrule
			\textbf{Ours (PMCR)} & \cmark & \textbf{\textcolor{red}{71.26}} & \textbf{\textcolor{red}{72.53}} & \textbf{\textcolor{red}{72.16}} & \textbf{\textcolor{red}{81.37}} & \textbf{\textcolor{red}{74.33}} & --- \\
			
			\bottomrule[1pt]
	\end{tabular}}
	\label{tab: Abdominal_CT} 
\end{table}

\begin{table}[t]
	\centering
	\setlength{\tabcolsep}{1.5mm}
	\caption{Comparisons (Dice score) on the Prostate-MRI \cite{Huang_2023_CVPR} dataset under the setting \#1. }
\resizebox{\linewidth}{!}{
		\begin{tabular}{l|c|cccc|cl}
			\toprule[1pt]
			\makecell[c]{Comparison Methods} & SSL & Fold1 & Fold2 & Fold3 & Fold4 & Avg. & Imp.  \\
			\midrule[1pt]
			LSLPNet \cite{9434008} & \xmark & 42.09 & 29.00& 32.49 &24.46& 32.01 & +33.53\\
			3dCANet \cite{LI2023102935}& \xmark & 59.36 & 60.38 & 45.73 &37.60& 50.77 & +14.77 \\
			VQ \cite{Huang_2023_CVPR}& \xmark & 63.77 & 61.32 &45.80 &38.11& 52.25 & +13.29 \\
			SSL-ALPNet \cite{10.1007/978-3-030-58526-6_45}& \cmark & 54.50 & 46.88 &66.38& 63.50 & 57.82 & +7.72 \\
			SSL-VQ \cite{Huang_2023_CVPR}& \cmark & 57.12 & 50.23 & 75.12 & 70.31 & 63.22 & +2.32 \\
			\midrule
			\textbf{Ours (PMCR)}& \cmark & \textbf{\textcolor{red}{59.31}} & \textbf{\textcolor{red}{53.05}} & \textbf{\textcolor{red}{77.28}} & \textbf{\textcolor{red}{72.51}} & \textbf{\textcolor{red}{65.54}} & -- \\
			\bottomrule[1pt]
	\end{tabular}}
	\label{tab: Prostate_MRI} 
\end{table}

\textbf{Evaluation Protocol:}
To evaluate segmentation performance of our model on 3D volumetric images, we follow traditional few-shot medical image segmentation works \cite{GUHAROY2020101587, Liu_2022_CVPR, Wang_2019_ICCV, 9709261, Huang_2023_CVPR}. Specifically, we use the protocol proposed in \cite{GUHAROY2020101587} to assign both the support and query sets. For each 3D volumetric image, we
equally divide its region-of-interest into $P$ equal-spaced chunks. Then we utilize the middle slice in each chunk from the support scan as reference to segment all slices in the corresponding query chunk. We set $P=3$ to conduct all comparison experiments.
In principle, the query and support scans are collected from different patients. After the support and query slices (2D images) are fed into the 2D segmentation model to obtain the 2D predictions, we follow \cite{GUHAROY2020101587, 9709261, Huang_2023_CVPR} to restack 2D segmentation results back into 3D volumes, and use Dice scores to measure overlapping between groundtruth and predicted segmentation masks. 
As introduced in \cite{10.1007/978-3-030-58526-6_45}, we employ two kinds of settings to validate the effectiveness of the proposed model on unseen novel medical classes:
\begin{itemize}
\item \textbf{The Setting \#1:}
In the training phase, testing classes are treated as background. This is a traditional experimental setting in the few-shot medical image segmentation. We select only one class for testing and the rest of the classes for training. 
\item \textbf{The Setting \#2:}
We remove the training images including any testing classes as background in the setting \#2. 
\end{itemize}

\begin{table}[t]
	\centering
	\setlength{\tabcolsep}{1.25mm}
	\caption{Ablation studies in terms of averaged Dice score on the Abdominal-MRI \cite{KAVUR2021101950} (top block) and Abdominal-CT \cite{10.1007/978-3-030-58526-6_45} (bottom block) under the setting 1.}
\resizebox{\linewidth}{!}{
		\begin{tabular}{c|l|cccc|c}
			\toprule[1pt]
			& \makecell[c]{Variants} & SSL& CRR & PCM & DCL & Avg. \\
			\midrule[1pt]
			\multirow{6}{*}{\rotatebox{90}{Abdominal-MRI}}  & Baseline & \xmark & \xmark & \xmark & \xmark & 38.62 \\
			& Baseline+SSL \cite{10.1007/978-3-030-58526-6_45} &\cmark & \xmark & \xmark & \xmark & 78.84 \\
			& Baseline+SSL+CRR & \cmark & \cmark & \xmark & \xmark & 83.35 \\
			& Baseline+SSL+CRR+PCM & \cmark & \cmark & \cmark & \xmark & 85.47 \\
			& Baseline+SSL+CRR+DCL & \cmark & \cmark & \xmark & \cmark & 84.63 \\
			& \textbf{Ours (PMCR)} & \cmark & \cmark &\cmark & \cmark & \textbf{\textcolor{red}{87.37}}  \\
			\midrule[1pt]
			\midrule[1pt]
			\multirow{6}{*}{\rotatebox{90}{Abdominal-CT}}  & Baseline & \xmark& \xmark & \xmark & \xmark &32.46  \\
			& Baseline+SSL \cite{10.1007/978-3-030-58526-6_45} &\cmark & \xmark & \xmark & \xmark & 73.35 \\
			& Baseline+SSL+CRR & \cmark & \cmark & \xmark & \xmark & 75.06 \\
			& Baseline+SSL+CRR+PCM & \cmark & \cmark & \cmark & \xmark & 76.21  \\
			& Baseline+SSL+CRR+DCL & \cmark & \cmark & \xmark & \cmark & 74.42 \\
			& \textbf{Ours (PMCR)} & \cmark & \cmark &\cmark & \cmark & \textbf{\textcolor{red}{77.23}}  \\
			
			\bottomrule[1pt]
	\end{tabular}}
	\label{tab: ablation_study_setting_1} 
\end{table}

\begin{table}[t]
	\centering
	\setlength{\tabcolsep}{1.25mm}
	\caption{Ablation studies in terms of averaged Dice score on  the Abdominal-MRI \cite{KAVUR2021101950} and Abdominal-CT \cite{10.1007/978-3-030-58526-6_45} datasets under the setting 2. }
\resizebox{\linewidth}{!}{
		\begin{tabular}{c|l|cccc|c}
			\toprule
			& \makecell[c]{Variants} & SSL& CRR & PCM & DCL & Avg. \\
			\midrule[1pt]
			\multirow{6}{*}{\rotatebox{90}{Abdominal-MRI}}  & Baseline & \xmark & \xmark & \xmark & \xmark & 35.17 \\
			& Baseline+SSL \cite{10.1007/978-3-030-58526-6_45} &\cmark & \xmark & \xmark & \xmark & 73.02 \\
			& Baseline+SSL+CRR & \cmark & \cmark & \xmark & \xmark & 76.64 \\
			& Baseline+SSL+CRR+PCM & \cmark & \cmark & \cmark & \xmark & 78.31 \\
			& Baseline+SSL+CRR+DCL & \cmark & \cmark & \xmark & \cmark & 79.11 \\
			& \textbf{Ours (PMCR)} & \cmark & \cmark &\cmark & \cmark & \textbf{\textcolor{red}{81.53}}  \\
			\midrule[1pt]
			\midrule[1pt]
			\multirow{6}{*}{\rotatebox{90}{Abdominal-CT}}  & Baseline & \xmark &\xmark & \xmark & \xmark &30.27  \\
			& Baseline+SSL \cite{10.1007/978-3-030-58526-6_45} &\cmark & \xmark & \xmark & \xmark & 63.02 \\
			& Baseline+SSL+CRR & \cmark & \cmark & \xmark & \xmark & 68.06  \\
			& Baseline+SSL+CRR+PCM & \cmark & \cmark & \cmark & \xmark & 72.34  \\
			& Baseline+SSL+CRR+DCL & \cmark & \cmark & \xmark & \cmark & 73.01 \\
			& \textbf{Ours (PMCR)} & \cmark & \cmark &\cmark & \cmark & \textbf{\textcolor{red}{74.33}}  \\
			\bottomrule
	\end{tabular}}
	\label{tab: ablation_study_setting_2} 
\end{table}

To facilitate direct performance comparison across different datasets and modalities for the same set of semantic classes, we create a standardized label set encompassing the liver, spleen, right kidney (RK) and left kidney (LK) for Abdominal-MRI \cite{KAVUR2021101950} and Abdominal-CT \cite{10.1007/978-3-030-58526-6_45}. The Cardiac-MRI \cite{8458220} dataset consists of three labels: left ventricle myocardium (LV-MYO), right ventricle (RV) and left ventricle blood pool (LV-BP). For Prostate-MRI \cite{Huang_2023_CVPR}, we follow \cite{Huang_2023_CVPR} to set 8 anatomical structure labels into four folds. These eight anatomical labels have complex inter-class relations and large intra-class variances. For fair comparisons, in this paper, we utilize five-fold cross-validation to perform comparison experiments.

\begin{figure*}[t]
	\centering
	\includegraphics[width=1.0\linewidth]{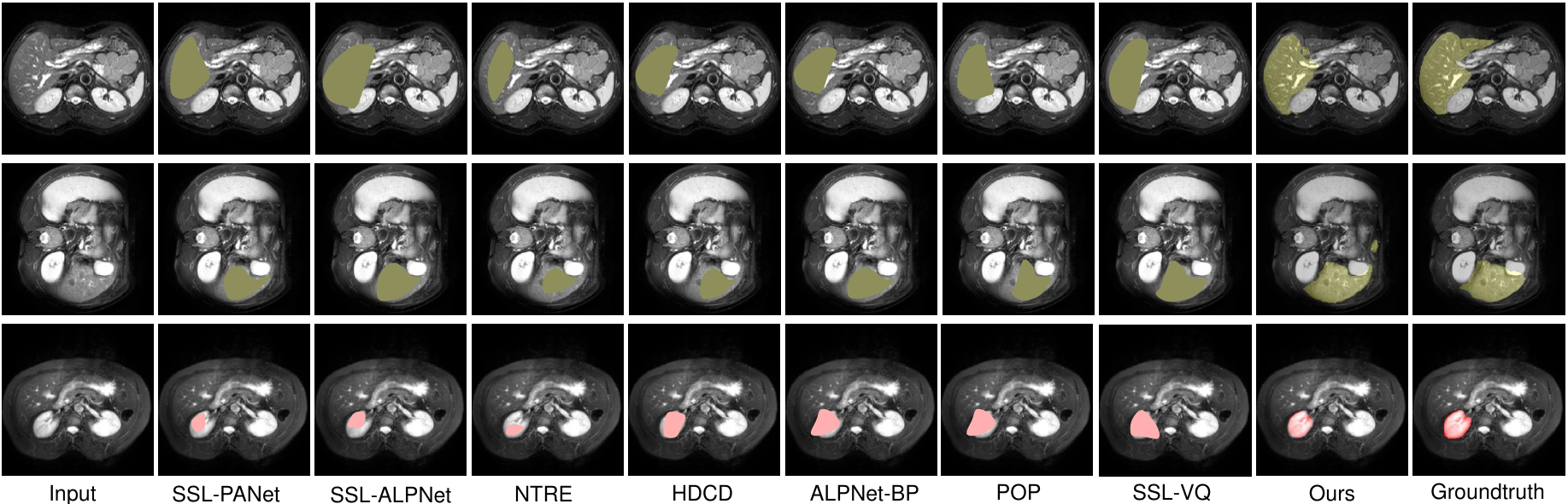}
 \vspace{-20pt}
	\caption{Exemplar segmentation results of the proposed model and some representative methods on Abdominal-MRI \cite{KAVUR2021101950} under the setting \#1.} 
	\label{fig:results_Abdominal_MRI_setting_1}
\end{figure*}

\begin{figure*}[t]
	\centering
	\includegraphics[width=1.0\linewidth]{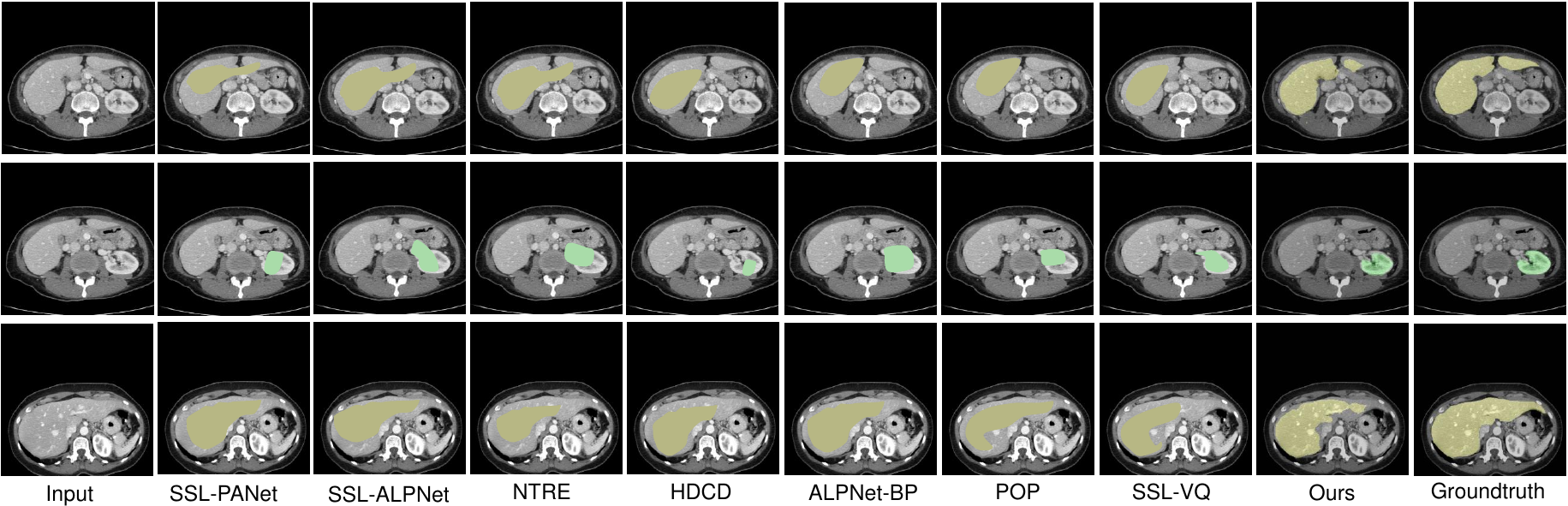}
        \vspace{-20pt}
	\caption{Exemplar segmentation results of the proposed model and some representative methods on Abdominal-CT \cite{10.1007/978-3-030-58526-6_45} under the setting \#1. }
	\label{fig:results_Abdominal_CT_setting_1}
\end{figure*}

\begin{table}[t]
	\centering
	\setlength{\tabcolsep}{1.5mm}
	\caption{Analysis of size of superpxiels ($G$) on Abdominal-MRI \cite{KAVUR2021101950} and Abdominal-CT \cite{10.1007/978-3-030-58526-6_45} datasets under the setting \#1.  }
		\scalebox{0.9}{
			\begin{tabular}{c|ccccc}
				\toprule[1pt]
				Size of Superpixels & 40 & 60 &80 & 100 & 120 \\
				\midrule[1pt]
				Abdominal-MRI & 86.02 & 86.14 & \textbf{\textcolor{red}{87.37}} & 87.04 & 86.27 \\
				Abdominal-CT  & 76.10 & 76.47 & \textbf{\textcolor{red}{77.23}} &75.86 & 75.14 \\
				\bottomrule[1pt]
		\end{tabular}}
		\label{tab: size_of_superpixels} 
		\vspace{-10pt}
	\end{table}

\subsection{Implementation Details}
In this paper, a fully convolutional ResNet-101 \cite{7780459} framework is utilized as a backbone network to get feature maps. For better medical image segmentation performance, inspired by \cite{Wang_2019_ICCV}, the ResNet-101 \cite{7780459} backbone is pretrained on the MS COCO dataset \cite{10.1007/978-3-319-10602-1_48}. 
We employ stochastic gradient descent (SGD) to optimize the objective in Eq.~\eqref{eq: overall_objective}.
Most of existing baseline few-shot medical image segmentation works \cite{GUHAROY2020101587, Liu_2022_CVPR, Wang_2019_ICCV, 9709261, Huang_2023_CVPR} perform 1-way 1-shot learning in the few-shot medical image segmentation task. Following these baseline methods which set the batch size as 1, we also set the exact same settings for fair comparisons. For the optimization, we set the learning rate as $0.001$, which is decayed by 0.95 after each $1,000$ iterations. 
The number of superpixel centroids $N_s$ is determined via $N_s = \min(\lfloor \frac{N_i}{G} \rfloor, N_s^{\mr{max}})$, where $N_i$ is the number of pixels within foreground mask, $G = 80$ denotes the averaged pixel number within each superpixel region. Motivated by \cite{10.1007/978-3-030-58526-6_45}, we set the maximum number of superpixel centroids as $N_s^{\mr{max}}=10$. We feed an image with size of $3\times 256\times 256$ to the ResNet-101 backbone \cite{7780459}, and obtain a latent feature with size of $256\times 32\times 32$. For the dimension of mapping matrices, we set $D_u=D_l=d=256$, and set the number of representative prototypes as $S=16$. In each training epoch, we use the Sinkhorn algorithm \cite{NIPS2013_af21d0c9} to optimize Eq.~\eqref{eq: optimization_OT}, and effectively obtain the optimal transportation matrix $\mathbf{T}^*\in\mathbb{R}^{S\times S}$. After getting the optimal $\mathbf{T}^*\in\mathbb{R}^{S\times S}$, we train the while networks via optimizing Eq.~\eqref{eq: overall_objective} in each training iteration. With well initialized centroids \cite{DBLP:journals/corr/Irving16}, the clustering algorithm SLIC \cite{6205760} only performs five iterations to obtain the final superpixel centroids. To reduce computation overhead, the number of multi-head self-attention is $H=1$.

\subsection{Comparison Experimental Results}
We present comparison experiments between the proposed PMCR model and some state-of-the-art (SOTA) few-shot segmentation methods \cite{Wang_2019_ICCV, 10.1007/978-3-030-58526-6_45, Huang_2023_CVPR, 9709261} to validate the superior performance of the proposed model. Tables~\ref{tab: cardiac_setting_1}--\ref{tab: Prostate_MRI} show the few-shot medical image segmentation performance on Cardiac-MRI \cite{8458220}, Abdominal-MRI \cite{KAVUR2021101950}, Abdominal-CT \cite{10.1007/978-3-030-58526-6_45}, and Prostate-MRI \cite{Huang_2023_CVPR} datasets under the settings \#1 and \#2. In the few-shot medical image segmentation, the SOTA baseline methods are \cite{10.1007/978-3-030-58526-6_45, Huang_2023_CVPR, 9709261}. When combining the self-supervised learning (SSL) proposed in \cite{10.1007/978-3-030-58526-6_45} with our model, the proposed model outperforms other methods by about $1.82\%\sim55.04\%$ Dice score under the setting \#1 and $2.27\%\sim62.42\%$ Dice score under the setting \#2. The proposed model performs better than some segmentation models \cite{10.1007/978-3-030-58526-6_45, GUHAROY2020101587, Wang_2019_ICCV} without SSL, since these methods cannot explore generalized features for base and novel medical classes via SSL. When compared with other comparison methods, the proposed PMCR model achieves large performance improvements in terms of the averaged Dice score. Besides, all evaluated approaches have better performance on the Abdominal-MRI than the Abdominal-CT, since there is a distinct contrast between different organs and surrounding tissues in Abdominal-MRI. 

Tables~\ref{tab: cardiac_setting_1}--\ref{tab: Prostate_MRI} also illustrate the better generalization performance of the proposed PMCR model on unseen medical classes. It implies the effectiveness of proposed class-relation reasoning (CRR) module to encode contextual information of query images via considering inbuilt inter-class relations between base and novel medical classes. Moreover, the proposed prototype correlation matching (PCM) module can improve the generation ability to determine accurate segmentation boundaries via exploring prototype-level correlation matching between support and query features to mitigate intra-class variations. 
As shown in Fig.~\ref{fig:results_Abdominal_MRI_setting_1}--\ref{fig:results_Abdominal_CT_setting_1}, we visualize some exemplar segmentation results of the PMCR model on Abdominal-MRI \cite{KAVUR2021101950} and Abdominal-CT \cite{10.1007/978-3-030-58526-6_45} datasets to further validate the effectiveness of proposed model. It illustrates that our model has the best segmentation performance that matches the groundtruth better than other competing methods~\cite{Huang_2023_CVPR, Wang_2019_ICCV}.

\begin{figure}[t]
	\centering
	\includegraphics[width=1.0\linewidth]
	{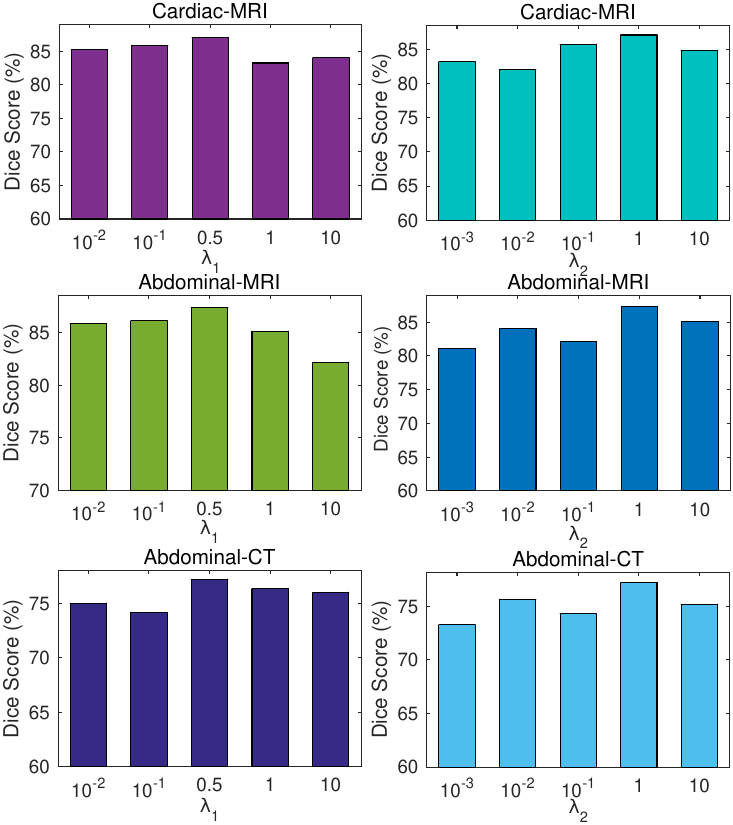}
	\vspace{-20pt}
	\caption{Analysis of hyper-parameters $\{\lambda_1, \lambda_2\}$ on Cardiac-MRI \cite{8458220} (top), Abdominal-MRI \cite{KAVUR2021101950} (middle) and Abdominal-CT \cite{10.1007/978-3-030-58526-6_45} (bottom) under the setting \#1. We analyze $\lambda_1$ when $\lambda_2=1.0$ and investigate $\lambda_2$ when $\lambda_1=0.5$.  }
	\label{fig: lambda_para}
\end{figure}

\begin{table}[t]
	\centering
	\setlength{\tabcolsep}{1.5mm}
	\caption{Analysis of different backbones in terms of averaged Dice score (\%) on the Abdominal-MRI \cite{KAVUR2021101950} under the setting \#1.  }
	\resizebox{1.0\linewidth}{!}{
			\begin{tabular}{l|c|ccc}
				\toprule[1pt]
				Comparison Methods & SSL& VGG-19 \cite{DBLP:journals/corr/SimonyanZ14a} & ResNet-101 \cite{7780459} & U-Net \cite{RFB15a} \\
				\midrule[1pt]
				SSL-PANet \cite{Wang_2019_ICCV} & \cmark & 58.41 & 63.17 & 65.26 \\
				SSL-ALPNet \cite{10.1007/978-3-030-58526-6_45} & \cmark & 74.18 & 78.84 & 81.26 \\
				RP-Net \cite{9711123} & \cmark & 75.61 & 79.76 & 82.36 \\
				ALPNet-BP \cite{9709261} & \cmark & 73.18 & 76.22 & 80.17 \\
				POP \cite{liu2023learning} &\cmark &71.33 & 74.77 & 79.62 \\
				SSL-VQ \cite{Huang_2023_CVPR} & \cmark & 80.15& 84.56 & 86.48 \\
				\midrule
				\textbf{Ours (PMCR)} & \cmark & \textbf{\textcolor{red}{82.66}}& \textbf{\textcolor{red}{87.37}}& \textbf{\textcolor{red}{89.04}}  \\
				\bottomrule[1pt]
		\end{tabular}}
		\label{tab: analysis_backbone} 
		\vspace{-10pt}
	\end{table}

\begin{table}[t]
\centering
\setlength{\tabcolsep}{1.5mm}
\caption{Analysis of mIoU (\%) on Abdominal-MRI \cite{KAVUR2021101950} and Abdominal-CT \cite{10.1007/978-3-030-58526-6_45} datasets under the setting \#1. }
\scalebox{0.9}{
\begin{tabular}{l|cc}
\toprule[1pt]
Comparison Methods &  Abdominal-MRI & Abdominal-CT \\
\midrule[1pt]
SE-Net \cite{GUHAROY2020101587} & 35.62 & 23.86 \\
ALPNet \cite{10.1007/978-3-030-58526-6_45} & 45.62 & 36.71 \\
\midrule
ALPNet-init \cite{10.1007/978-3-030-58526-6_45} &  60.16 & 58.77 \\
SSL-ALPNet \cite{10.1007/978-3-030-58526-6_45}& 72.37 & 66.14 \\
ALPNet-BP \cite{9709261} & 70.66 & 67.83\\
POP \cite{liu2023learning} & 68.52 & 64.91\\
SSL-VQ \cite{Huang_2023_CVPR} &78.30 & 69.96 \\
\midrule
\textbf{Ours (PMCR)} & \textbf{\textcolor{red}{82.07}} & \textbf{\textcolor{red}{71.85}} \\
\bottomrule[1pt]
\end{tabular}}
\label{tab: comparison_mIoU} 
\end{table}	

\begin{figure}[t]
	\centering
	\includegraphics[width=1.0\linewidth]
	{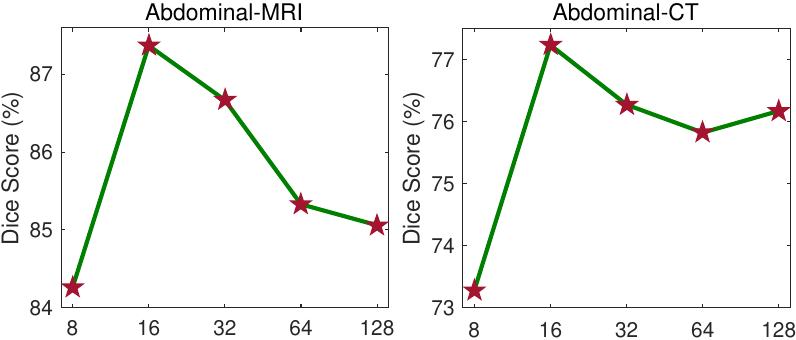}
	\vspace{-20pt}
	\caption{Analysis of number of prototypes on the Abdominal-MRI dataset \cite{KAVUR2021101950} (left) and Abdominal-CT dataset \cite{10.1007/978-3-030-58526-6_45} (right) under the setting \#1 in the few-shot medical image segmentation.} 
	\label{fig:prototype_MRI_CT}
\end{figure}

\subsection{Ablation Study}
Tables~\ref{tab: ablation_study_setting_1}--\ref{tab: ablation_study_setting_2} show the ablation experiments of the proposed model on Abdominal-MRI \cite{KAVUR2021101950} and Abdominal-CT \cite{10.1007/978-3-030-58526-6_45} datasets under the setting \#1 and \#2, respectively. For simplification, the proposed class-relation reasoning module, prototype correlation matching module, and Dice loss $\mathcal{L}_{\mr{dc}}$ are abbreviated as CRR, PCM and DCL, respectively. More importantly, we also need to introduce ablation experiments about self-supervised learning (SSL) proposed in \cite{10.1007/978-3-030-58526-6_45}, since we combine the SSL proposed in \cite{10.1007/978-3-030-58526-6_45} with our proposed model to conduct comparison experiments. The baseline represents the performance of our model without using the SSL, CRR, PCM and DCL modules. When compared with Ours, all variants degrade the performance about $1.32\%\sim11.31\%$ in terms of averaged Dice score. It implies the effectiveness and superiority of each proposed module. When we neglect one of the proposed modules, the large performance decrease on unseen medical classes illustrates the complementarity of proposed modules in the few-shot medical image segmentation task. The proposed CRR module could effectively exploit inter-class relations to encode semantic features of unseen novel medical classes. 
As shown in Table~\ref{tab: size_of_superpixels}, we analyze the effect of the size of superpixels $G$, and observe that our model has the best performance when setting $G=80$. Both an excess and a shortage of superpixels may introduce noise information, making it impossible to extract accurate inter-class relations and resulting in decreased segmentation performance. 
Furthermore, when adding the PCM module to the baseline, it increases performance largely by exploring prototype-level correlation matching between support and query features to mitigate intra-class variations. Meanwhile, the Dice loss $\mathcal{L}_{\mr{dc}}$ is also an essential objective function to perform few-shot medical image segmentation task, which is effective in distinguishing foreground regions from the surrounding backgrounds. When compared with the baseline method \cite{10.1007/978-3-030-58526-6_45} with the SSL, the performance of our model improves significantly in terms of averaged Dice score. 
Overall, these results illustrate the effectiveness of each module.

\begin{table}[t]
	\centering
	\setlength{\tabcolsep}{1.5mm}
	\caption{Analysis of representative prototypes produced by SVD on Abdominal-MRI \cite{KAVUR2021101950} dataset under the setting \#1 (top block) and the setting \#2 (bottom block). } 
		\scalebox{0.9}{
			\begin{tabular}{l|cccc|c}
				\toprule[1pt]
				\makecell[c]{Variants} & LK & RK & Spleen & Liver & Avg. \\
				\midrule[1pt]
				Ours-w/ MAP& 80.76 & 81.69&90.31 & 86.26 & 84.76 \\ 
				\textbf{Ours (PMCR)} & \textbf{\textcolor{red}{83.27}} & \textbf{\textcolor{red}{84.04}} & \textbf{\textcolor{red}{92.33}} & \textbf{\textcolor{red}{89.82}} & \textbf{\textcolor{red}{87.37}} \\
				\midrule[1pt]
				\midrule[1pt]
				Ours-w/ MAP&76.14 &81.28 &84.73 &73.06 & 78.80\\
				\textbf{Ours (PMCR)} & \textbf{\textcolor{red}{78.89}} & \textbf{\textcolor{red}{84.11}} & \textbf{\textcolor{red}{87.64}} & \textbf{\textcolor{red}{75.49}} & \textbf{\textcolor{red}{81.53}} \\
				\bottomrule[1pt]
		\end{tabular}}
		\label{tab: analysis_SVD} 
\end{table}
\subsection{Analysis of Hyper-Parameters $\lambda_1, \lambda_2$}
As presented in Fig.~\ref{fig: lambda_para}, we analyze the effect of hyper-parameters $\{\lambda_1, \lambda_2\}$ in Eq.~\eqref{eq: overall_objective} on the Cardiac-MRI \cite{8458220}, Abdominal-MRI \cite{KAVUR2021101950} and Abdominal-CT \cite{10.1007/978-3-030-58526-6_45} datasets. 
To determine the values of hyper-parameters $\{\lambda_1, \lambda_2\}$, we follow the traditional few-shot medical image work \cite{9711123} to set $\lambda_2$ as 1.0, and fix $\lambda_2=1.0$ to select the best $\lambda_1$ from the set $\{0.1, 0.5, 1.0, 3.0, 5.0\}$ by evaluating which one yields the best segmentation performance. Then we can determine the best values of $\lambda_1$ and $\lambda_2$ as 0.5 and 1.0 via only a few parameter experiments. In order to further investigate the generalization of our model over a wide selection range of parameters, we select the values of $\lambda_1, \lambda_2$ in a wide range of $\{10^{-3}, 10^{-2}, 10^{-1}, 1.0, 10\}$ and $\{10^{-2}, 10^{-1}, 0.5, 1.0, 10\}$, respectively, to evaluate the averaged Dice score of proposed model on the Cardiac-MRI \cite{8458220}, Abdominal-MRI \cite{KAVUR2021101950} and Abdominal-CT \cite{10.1007/978-3-030-58526-6_45} datasets under the setting \#1. As shown in Fig.~\ref{fig: lambda_para}, when $\lambda_2=1.0$, we investigate the performance of $\lambda_1=\{10^{-2}, 10^{-1}, 0.5, 1.0, 10\}$; we also show the results of $\lambda_2=\{10^{-3}, 10^{-2}, 10^{-1}, 1.0, 10\}$ when setting $\lambda_1=0.5$. From the shown results, we observe that the PMCR model has stable segmentation performance for a wide selection range of $\{\lambda_1, \lambda_2\}$, which could well generalize to unseen medical classes to improve few-shot segmentation performance. Moreover, the proposed model has the best generalization ability to segment unseen novel medical classes when setting $\lambda_1=0.5, \lambda_2=1.0$ under different experimental settings.

\begin{figure}[t]
	\centering
	\includegraphics[width=0.7\linewidth]{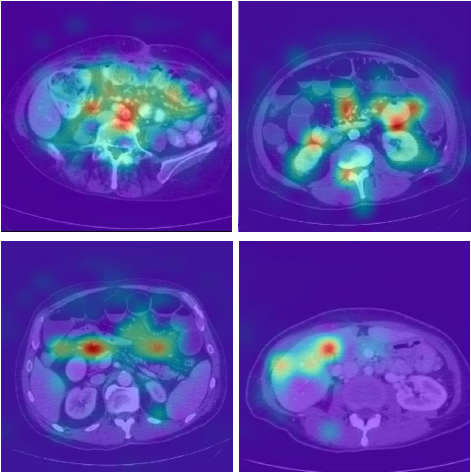}
	\vspace{-5pt}
	\caption{Visualization of attention map on Abdominal-MRI \cite{KAVUR2021101950} (top) and Abdominal-CT \cite{10.1007/978-3-030-58526-6_45} (bottom) under the setting \#1. }
	\label{fig: attention_map}
	\vspace{-10pt}
\end{figure}

\subsection{Effect of Different Backbones}
As presented in Table~\ref{tab: analysis_backbone}, we analyze the effect of different backbones on the performance of few-shot medical image segmentation. From the averaged Dice scores in Table~\ref{tab: analysis_backbone}, we notice that the proposed model has better segmentation performance than other comparison methods in the few-shot medical image segmentation task, even though we use different backbones to extract the local features. It illustrates that the proposed model does not have a significant dependence on the choices of backbones to perform few-shot medical image segmentation. Our model surpasses the existing state-of-the-art few-shot methods by a large margin of averaged Dice scores. To accurately segment unseen new medical classes, we can explore inherent relations among base and novel medical classes via the class-relation reasoning module and effectively address the false pixel matches caused by large intra-class variations via the prototype correlation matching module.

\subsection{Analysis of Evaluation Metrics}
In order to further verify the effectiveness of proposed model, we use mean Intersection over Union (mIoU) that is commonly-used metric in the few-shot segmentation tasks of natural images to evaluate the performance. As shown in Table~\ref{tab: comparison_mIoU}, the proposed PMCR model significantly outperforms other methods by a large margin of mIoU. 
Considering the comparison experiments in Tables~\ref{tab: cardiac_setting_1}--\ref{tab: Prostate_MRI}, we conclude that the proposed model has the best few-shot segmentation performance than other comparison methods, under different evaluation metrics and settings. Such improvement validates the effectiveness of our model in addressing the false pixel matches brought by large intra-class variations and exploring inter-class relations among base and novel medical classes.

\subsection{Analysis of The PCM Module}
In this subsection, we aim to investigate the effect of the number of prototypes in the PCM module. As shown in Fig.~\ref{fig:prototype_MRI_CT}, we set $S=\{8, 16, 24, 32, 64, 128\}$ to show the performance of proposed PMCR model on Abdominal-MRI \cite{KAVUR2021101950} and Abdominal-CT \cite{10.1007/978-3-030-58526-6_45} datasets under the setting \#1. When $S=16$, the proposed PCM model has the best performance. It illustrates that these prototypes initialized by Singular Value Decomposition (SVD) can embody diverse visual information of different appearances and shapes from support and query images. When we explore prototype-level correlation matching between support and query features via the optimal transport, the false pairwise pixel correlation matches brought by large intra-class variations can be effectively mitigated to improve few-shot medical segmentation.

\begin{figure}[t]
	\centering
	\includegraphics[width=1.0\linewidth]{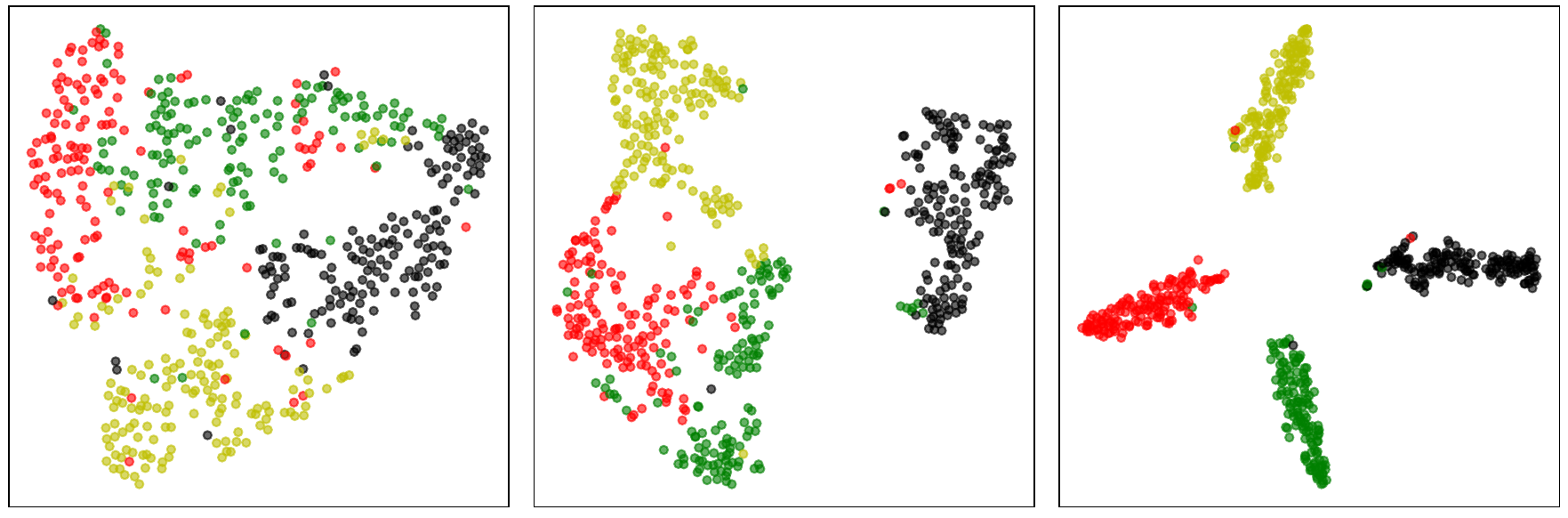}
	\vspace{-20pt}
	\caption{TSNE visualization \cite{inbook} of latent features extracted by our model without training (left), our model without PCM (middle) and our model with PCM (right) on Abdominal-MRI \cite{KAVUR2021101950} under the setting \#1. }
	\label{fig: cluster_MRI1}
\end{figure}

\begin{figure}[t]
	\centering
	\includegraphics[width=1.0\linewidth]{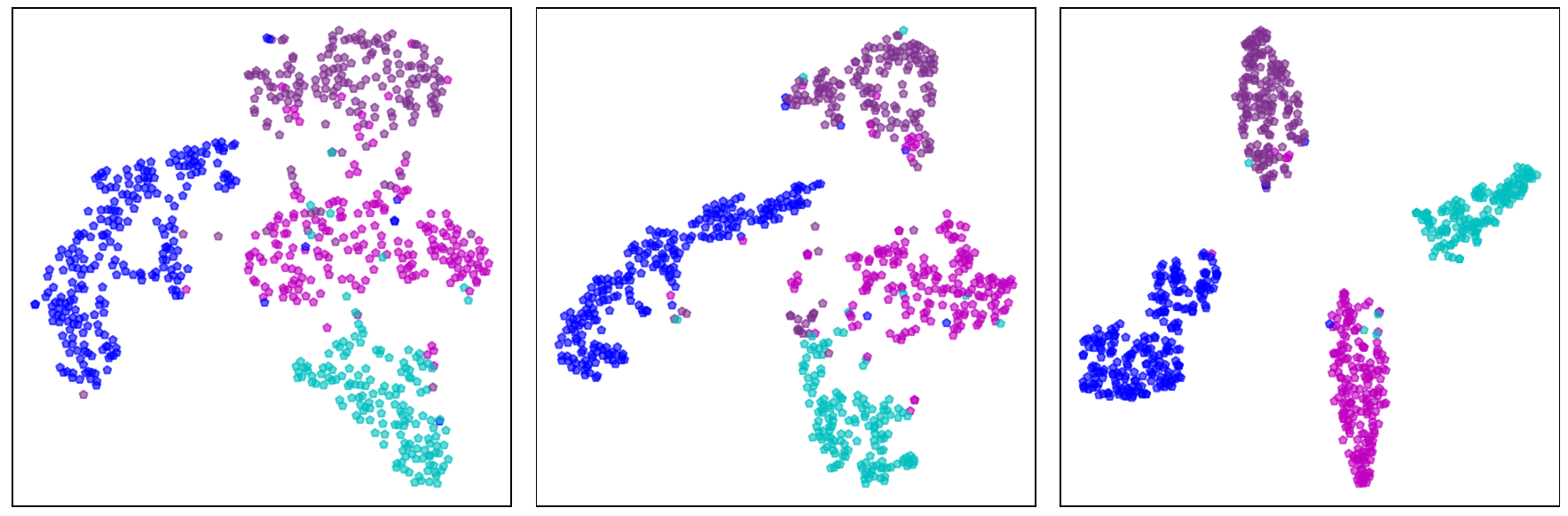}
	\vspace{-20pt}
	\caption{TSNE visualization \cite{inbook} of latent features extracted by our model without training (left), our model without PCM (middle) and our model with PCM (right) on Abdominal-CT \cite{10.1007/978-3-030-58526-6_45} under the setting \#1.  }
	\label{fig: cluster_CT1}
	\vspace{-10pt}
\end{figure}

\begin{table}[t]
\centering
\setlength{\tabcolsep}{1.5mm}
\caption{Analysis of the class-relation reasoning (CRR) module on Abdominal-CT \cite{10.1007/978-3-030-58526-6_45} dataset under the setting \#1 (top block) and the setting \#2 (bottom block). } 
\scalebox{0.9}{
	\begin{tabular}{l|c|cccc|c}
		\toprule[1pt]
		\makecell[c]{Variants} &SSL & LK & RK & Spleen & Liver & Avg. \\
		\midrule[1pt]
		HDCD \cite{peng2023cvpr} & \cmark & 70.36 & 71.22 & 74.19 & 73.82 & 72.40 \\
		Ours-w/ TF & \cmark& 72.65 & 73.86 & 74.30 & 76.87 & 74.42 \\ 
		\textbf{Ours (PMCR)} & \cmark & \textbf{\textcolor{red}{77.16}} & \textbf{\textcolor{red}{75.27}} & \textbf{\textcolor{red}{74.34}} & \textbf{\textcolor{red}{82.16}} & \textbf{\textcolor{red}{77.23}}  \\
		\midrule[1pt]
		\midrule[1pt]
		HDCD \cite{peng2023cvpr} & \cmark & 67.46 & 68.73 & 70.12 & 80.04 & 71.59 \\
		Ours-w/ TF & \cmark &68.93 & 70.14 & 71.26 & 80.42 & 72.69 \\
		\textbf{Ours (PMCR)} & \cmark & \textbf{\textcolor{red}{71.26}} & \textbf{\textcolor{red}{72.53}} & \textbf{\textcolor{red}{72.16}} & \textbf{\textcolor{red}{81.37}} & \textbf{\textcolor{red}{74.33}} \\
		\bottomrule[1pt]
\end{tabular}}
\label{tab: analysis_CRR} 
\end{table}

To further analyze the effectiveness of SVD, as shown in Table~\ref{tab: analysis_SVD}, we use the global class prototypes generated by the masked average pooling (MAP) to replace the prototypes produced by SVD, and denote this variant as Ours-w/ MAP. The comparison experiments in Table~\ref{tab: analysis_SVD} illustrate that the prototypes generated via SVD are more effective to address false pixel matches brought by large intra-class variations, compared with global prototypes produced via MAP. As shown in Fig.~\ref{fig: attention_map}, we have visualized some attention maps by mapping them into the input images. These visualization results illustrate that the proposed model can effectively capture important local contextual information of query images to address the few-shot medical image segmentation task. 
More importantly, as shown in Figs.~\ref{fig: cluster_MRI1}--\ref{fig: cluster_CT1}, we introduce some t-Distributed Stochastic Neighbor Embedding (TSNE) \cite{inbook} visualization of latent features extracted by our model without training (left), our model without the proposed PCM module (middle), and our model with the PCM module (right) on Abdominal-MRI \cite{KAVUR2021101950} and Abdominal-CT \cite{10.1007/978-3-030-58526-6_45} under the setting \#1. It verifies the effectiveness of PCM module to address false pairwise pixel correlation matches brought by large intra-class variations.

\begin{table}[t]
	\centering
	\setlength{\tabcolsep}{1.5mm}
	\caption{Comparisons of computation costs on the Cardiac-MRI \cite{8458220} dataset under the setting \#1. }
	\scalebox{0.9}{
		\begin{tabular}{l|c|cc}
			\toprule[1pt]
			\makecell[c]{Comparison Methods} & SSL & Training Time & Testing Time  \\
			\midrule[1pt]
			SSL-PANet \cite{Wang_2019_ICCV} & \cmark & 0.83s & 0.24s \\
			SSL-ALPNet \cite{10.1007/978-3-030-58526-6_45} & \cmark & 0.42s & 0.21s \\
			RP-Net \cite{9711123} & \cmark & 1.83s & 0.65s \\
			HDCD \cite{peng2023cvpr}  & \cmark & 1.33s & 0.36s \\
			ALPNet-BP \cite{9709261} & \cmark & 0.62s & 0.23s \\
			POP \cite{liu2023learning} & \cmark & 1.51s & 0.43s \\
			SSL-VQ \cite{Huang_2023_CVPR} & \cmark & 0.86s & 0.33s \\
			Ours-w/ TF & \cmark & 1.98s & 0.72s \\
			\midrule
			\textbf{Ours (PMCR)} & \cmark & 1.47s & 0.38s \\
			\bottomrule[1pt]
	\end{tabular}}
	\label{tab: computation_cost} 
\end{table}

\begin{table}[t]
	\centering
	\setlength{\tabcolsep}{1.5mm}
	\caption{Comparisons (Dice score) on the Cardiac-MRI \cite{8458220} dataset under the setting \#1 when we perform 1-way 5-shot medical image segmentation task. }
	\resizebox{1.0\linewidth}{!}{
		\begin{tabular}{l|c|ccc|cl}
			\toprule[1pt]
			\makecell[c]{Comparison Methods} & SSL & LV-BP & LV-MYO & RV & Avg. & Imp.  \\
			\midrule[1pt]
			Vanilla PANet \cite{Wang_2019_ICCV}  & \xmark & 60.69 & 39.44 &41.66 & 47.26 & +37.01 \\ 
			ALPNet \cite{10.1007/978-3-030-58526-6_45} & \xmark &82.65 &52.61& 69.13 &68.14 & +19.08 \\
			\midrule
			SSL-PANet \cite{Wang_2019_ICCV} & \cmark &70.62& 46.03& 67.16 &61.27 & +25.95 \\
			
			ALPNet-BP \cite{9709261} & \cmark & 86.89 & 72.14 & 85.95 & 81.66 & +5.56\\
			SSL-VQ \cite{Huang_2023_CVPR} & \cmark & 89.36 & 78.06 & 85.38 & 84.27 & +2.95 \\
			\midrule
			\textbf{Ours (PMCR)} & \cmark & \textbf{\textcolor{red}{91.10}} & \textbf{\textcolor{red}{80.14}} & \textbf{\textcolor{red}{90.42}} & \textbf{\textcolor{red}{87.22}} & -- \\
			\bottomrule[1pt]
	\end{tabular}}
	\label{tab: cardiac_setting_1_5shot} 
	\vspace{-10pt}
\end{table}

\subsection{Analysis of The CRR Module}
As shown in Table~\ref{tab: analysis_CRR}, we analyze the effectiveness of our proposed class-relation reasoning (CRR) module. 
In Table~\ref{tab: analysis_CRR}, we replace the proposed CRR module with the Transformer-Base backbone \cite{dosovitskiy2021an} to capture inter-class relations between query and support features, and denote this variant as Ours-w/ TF. When compared with Ours, the performance of Ours-w/ TF decrease significantly, since the variant Ours-w/ TF cannot determine the contributions of inter-class relations to the few-shot medical image segmentation tasks and adaptively refine semantic encoding of query images based on their different contributions. Therefore, some noisy inter-class relations caused by large intra-class verifies will significantly affect the performance of Ours-w/ TF on the few-shot medical image segmentation tasks. In contrast, our model uses inter-class relations to construct convolution kernel attention (\emph{i.e.}, contribution weights) via the proposed CRR module, and then propagate inter-class relations into the semantic encoding of local query features by masking the original convolution kernel with the convolution kernel attention. Compared with Ours-w/ TF, this strategy can address the negative influence of noisy inter-class relations by masking convolution kernel attention with different contribution weights of inter-class relations, thereby significantly outperforming the performance of Ours-w/ TF. Moreover, we introduce the complexity analysis of Our-w/ TF in Table~\ref{tab: computation_cost}. In Tables~\ref{tab: analysis_CRR}--\ref{tab: computation_cost}, compared to our model with ResNet-101 as the backbone, Ours-w TF with Transformer-Base as the backbone has higher computational cost but lower performance due to its limitations in tackling noisy inter-class relations. It validates the effectiveness of our CRR module to explore inter-class relations.

\subsection{Analysis of Computation Costs}
To analyze the efficiency of proposed model, as shown in Table~\ref{tab: computation_cost}, we introduce comparisons of computation costs between our model and comparison methods. All comparison experiments are conducted on a single Nvidia RTX 3080Ti GPU under the Ubuntu 16.01 system. From Table~\ref{tab: computation_cost}, we conclude that our model is efficient to perform few-shot medical image segmentation task, when compared with other methods. There are some reasons for our model to be efficient: 1) We follow baseline few-shot medical image segmentation methods \cite{9709261, 10.1007/978-3-030-58526-6_45} to only perform 1-way 1-shot learning. Our model is efficient under the setting of 1-way 1-shot.
2) In the PCM module, the number of foreground pixels $N_i$ is much smaller than $H_lW_l$: $N_i \ll H_lW_l$, and we set $H_l=W_l=32$. Besides, $\mathbf{F}^q$ and $\mathbf{F}_i^s$ have a very small feature dimension ($D_l=256$). We follow efficient SVD algorithm \cite{article_Halko2011} to efficiently obtain $\mathbf{U}, \Sigma_s$ and $\mathbf{V}$, where $S=16$ is also very small. It costs about 500ms to converge under a single Nvidia RTX 3080Ti GPU. 3) In the CRR module, we follow \cite{DBLP:journals/corr/Irving16} to obtain well initialized centroids. The clustering algorithm SLIC \cite{6205760} only perform five iterations (about 250ms) to obtain the final superpixel centroids. 4) To reduce computation overhead, the number of multi-head self-attention is set as $H=1$. In conclusion, when compared with the large performance improvement of our model (see Tables~\ref{tab: cardiac_setting_1}--\ref{tab: Prostate_MRI}), the additional computation costs can be negligible, and is acceptable in clinical diagnosis.

\subsection{Analysis of Multiple-Shot Scenarios}
This subsection analyze the performance of multiple-shot scenarios. As shown in Table~\ref{tab: cardiac_setting_1_5shot}, we conduct comparison experiments (Dice score) on the Cardiac-MRI \cite{8458220} dataset under the setting \#1 when we perform 1-way 5-shot medical image segmentation task. In Table~\ref{tab: cardiac_setting_1_5shot}, our model significantly outperforms other representative comparison methods by a large margin in terms of Dice score. Such performance improvement verifies the effectiveness of our model in exploring inter-class relations among base and novel classes. Besides, compared with the 1-shot results, our model performs better under the 5-shot scenarios. It illustrates that more samples of unseen classes can provide more supervision to address false pixel matches brought by large intra-class variance. 

\section{Conclusion} 
In this paper, we propose a prototype correlation matching and class-relation reasoning (PMCR) model to mitigate false pixel correlation matches caused by large intra-class variations and explore inter-class relations among different medical classes. To be specific, we propose a prototype correlation matching (PCM) module to tackle false pixel correlation matches brought by large intra-class variations. The proposed PCM module can mine representative prototypes to characterize diverse visual information of different appearances, and capture prototype-level rather than pixel-level correlation matching between support and query features via the optimal transport algorithm to mitigate false matches caused by intra-class variations. Furthermore, we develop a class-relation reasoning (CRR) module to explore inter-class relations, and exploit such relations between base and novel classes to segment unseen medical objects via refining semantic encoding of local query features. Extensive experiments on some benchmark medical MRI and CT datasets are performed to validate the superior few-shot medical segmentation performance of the proposed PMCR model. 
In the future, we will extend the proposed model to other challenging medical image segmentation tasks, such as federated learning-based few-shot segmentation \cite{Dong_2022_CVPR, Dong_2023_CVPR} and few-shot continual segmentation \cite{Gu_2023_ICCV}.  Moreover, we will consider different modalities of medical diseases to improve the accuracy and efficiency of clinical diagnosis.


\bibliographystyle{IEEEtran}
\bibliography{FSLMed}
\end{document}